\pdfoutput=1

\documentclass[usenames,dvipsnames]{article}
\usepackage{iclr2023arxiv,times}
\usepackage{hyperref}
\usepackage{url}

\usepackage{latexsym}

\usepackage[T1]{fontenc}

\usepackage[utf8]{inputenc}

\newcommand{\msmarco}{MS MARCO}

\newcommand{\nq}{NQ}
\newcommand{\hotpotqa}{HotpotQA}
\newcommand{\fiqa}{FiQA}

\newcommand{\arguana}{ArguAna}
\newcommand{\touche}{Touche-2020}

\newcommand{\dbpedia}{Dbpedia-Entity}

\newcommand{\fever}{FEVER}
\newcommand{\climatefever}{Climate-FEVER}

\newcommand{\flan}{FLAN}
\newcommand{\nqqgen}{NQ-QGen}

\usepackage{microtype}

\usepackage{array}
\usepackage{amsmath}
\usepackage{amsfonts}
\usepackage{graphicx}
\usepackage{bm}
\usepackage{bbm}
\usepackage{booktabs}
\usepackage{xcolor} %
\usepackage{subcaption}
\usepackage{booktabs} %
\usepackage{tikz} %
\usepackage{tikz-layers} %
\usepackage{xcolor} %
\usepackage{enumitem}
\usepackage{footnote}
\usepackage{mathtools}
\usepackage{multicol}
\usepackage{latexsym}
\usepackage{multirow}
\usepackage{natbib}
\usepackage{outlines}
\usepackage{pifont}%
\usepackage{subcaption}
\usepackage{soul}
\usepackage{tabularx}
\usepackage{tabulary}
\usepackage{subcaption}

\usepackage{url}
\usepackage{xcolor}
\usepackage{xspace}
\usepackage[utf8]{inputenc}
\usepackage{hyperref}
\usepackage{cleveref}
\usepackage{diagbox}
\usepackage{tablefootnote}
\usepackage{pgfplots} 
\usepackage{wrapfig}

\usepackage{booktabs}
\usepackage{amsfonts}
\usepackage{amsmath}
\usepackage{amssymb}%

\usepackage{enumitem}
\usepackage{etoolbox}

\usepackage{tikz}
\usetikzlibrary{arrows.meta}
\usetikzlibrary{patterns}

\tikzset{
  barlabels/.style={font=\footnotesize\sffamily},
  declare function={
    barheight=5pt;
  }
}

\newcommand{\cmark}{\ding{51}}%

\crefformat{section}{(\S#2#1#3)} %

\usepackage{header}

\usepackage{amsmath,amsfonts,bm}

\def\eqref#1{equation~\ref{#1}}

\def\1{\bm{1}}

\DeclareMathAlphabet{\mathsfit}{\encodingdefault}{\sfdefault}{m}{sl}
\SetMathAlphabet{\mathsfit}{bold}{\encodingdefault}{\sfdefault}{bx}{n}

\newcolumntype{R}{>{\raggedleft\arraybackslash}p{3em}}
\newcolumntype{C}{>{\centering\arraybackslash}p{3em}}
\newcommand{\nlp}[1]{\texttt{\small #1}}
\newcommand{\fullname}{Prompt-base Query Generation for Retriever\xspace}

\newcommand{\name}{{\small \textsc{Promptagator}}\xspace}
\newcommand{\namep}{{\small \textsc{Promptagator++}}\xspace}

\usepackage{scalerel}

\title{{\bf \textsc{Promptagator}\includegraphics[width=1.1em]{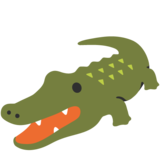}}: Few-shot Dense Retrieval From 8 Examples}

\author{Zhuyun Dai$^{*\dagger}$, Vincent Y. Zhao$^{*\dagger}$, Ji Ma$^{*\dagger}$, Yi Luan$^{*\dagger}$, Jianmo Ni, Jing Lu, Anton Bakalov, \\\textbf{Kelvin Guu, Keith B. Hall and Ming-Wei Chang$^{\dagger}$} \\
{\rm Google Research}\\
\texttt{\{zhuyundai, vzhao, maji,
luanyi, mingweichang\}@google.com}\\
{\footnotesize ${^*}$equal contributions~~${^\dagger}$corresponding authors}\\
}

\iclrfinalcopy
\begin{document}
\maketitle
\begin{abstract}

Much recent research on information retrieval has focused on how to transfer from one task (typically with abundant supervised data) to various other tasks where supervision is limited, with the implicit assumption that it is possible to generalize from one task to all the rest. However, this overlooks the fact that there are many diverse and unique retrieval tasks, each targeting different search intents, queries, and search domains. In this paper, we suggest to work on {\em Few-shot Dense Retrieval}, a setting where each task comes with a short description and a few examples. To amplify the power of a few examples, we propose \fullname (\name\includegraphics[width=1.0em]{tables_and_figures/crocodile_1f40a.png}), which leverages large language models (LLM) as a few-shot query generator, and creates task-specific retrievers based on the generated data. Powered by LLM's generalization ability,  \name makes it possible to create task-specific end-to-end retrievers solely based on a few examples {without} using Natural Questions~\citep{kwiatkowski2019natural} or \msmarco~\citep{nguyen2016msmarco} to train %
dual encoders. Surprisingly, LLM prompting with no more than 8 examples allows dual encoders to  outperform heavily engineered models trained on \msmarco~ like ColBERT v2~\citep{Santhanam2022ColBERTv2EA}
by more than 1.2 nDCG on average on 11 retrieval sets. Further training standard-size re-rankers using the {\em same} generated data yields another 5.0 point nDCG improvement. Our studies determine that query generation can be far more effective than previously observed, especially when a small amount of task-specific knowledge is given. 

\end{abstract}

\section{Introduction}
Recently, major progress has been made on neural retrieval models
such as dual encoders, which can retrieve knowledge from a large collection of documents containing millions to billions of passages
~\citep{yih-etal-2011-learning,lee-etal-2019-latent,dense-passage-qa}. However, \cite{thakur2021beir} recently proposed the BEIR heterogeneous retrieval benchmark, and showed that it is still difficult for neural retrievers to perform well on a wide variety of retrieval tasks that lack dedicated training data. 
Thus, previous approaches focus on transferring knowledge from question answering (QA) datasets such as
\msmarco{}~\citep{nguyen2016msmarco}. To best transfer from QA datasets,  expressive retrievers
are developed that allow fine-grained token-level interaction such as ColBERT~\citep{Colbert,Santhanam2022ColBERTv2EA} and SPLADE~\citep{spladev2}
but with higher inference cost.
Data augmentation via synthetic question generation has previously been explored~\citep{ma-etal-2021-zero,siamak20}, but these question generators are typically only trained on popular QA datasets.

\begin{figure}
    \centering
    \includegraphics[width=.95\linewidth]{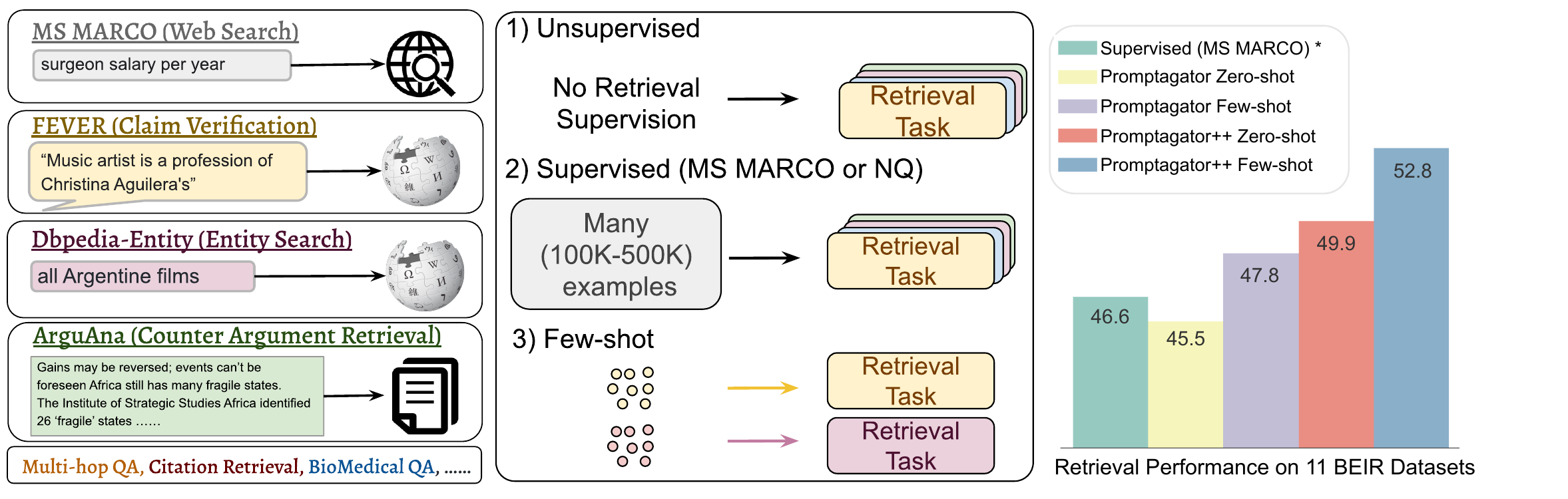}
    \caption{Few-shot retrieval with \name. \textbf{Left (a)}:  Retrieval tasks from BEIR differ in query distribution, retrieval corpus, and search intents. \textbf{Middle (b)}:  Most prior work uses supervised setting (2) which trains model on a large QA retrieval datasets and transfer to other retrieval tasks. \textbf{Right (c)}: Few-shot \name{} performance. Average nDCG@10 on 11 datasets from BEIR from our \name models and previously  \msmarco-supervised models (SPLADE v2).}
    \label{fig:fewshot}
\end{figure}

We argue that it is hard to expect models based on one or two QA datasets to perform well across different retrieval tasks. First, different retrieval tasks have very different \emph{search intents}; in other words, different definitions of ``relevance''. For example, as illustrated in Figure~\ref{fig:fewshot}(a),  both \dbpedia~\citep{hasibi2017dbpedia} and \fever~\citep{thorne-etal-2018-fever} are tasks to retrieve documents from Wikipedia. \dbpedia~ is a task to retrieve entities that are mentioned in the query, while \fever~ is a task to find evidence that either supports or refutes a given statement. Which document is relevant to the query can be very different from one task to another task even if they share the same domain. Moreover, different tasks have distinct distributions of queries even when their search intents are similar. For example, in the BEIR benchmark, queries in \hotpotqa~\citep{yang-etal-2018-hotpotqa} are long compositional questions, while queries in \fiqa~\citep{maia201818} are short financial questions. %

In this paper, we advocate to work on the setting of {\em Few-shot Retrieval} for diverse retrieval  \cref{sec:task}, where each task comes with a short description and a few annotated examples to clearly illustrate the search intents. Given that only a few examples are available, we propose {\fullname} (\name{}) \cref{sec:model} which aims to resolve the data scarcity issue while retaining the efficiency of a small dual encoder, by harnessing the power of large language models (LLM) such as FLAN~\citep{flan}. \name combines prompting with LLMs as a query generator without fine-tuning \cref{sec:prompt_qgen}, and can generate good queries  with minimal supervision -- shown in Figure~\ref{fig:fewshot}(b), it solely relies on a few supervised examples from the target task without using annotated query-document pairs from Natural Questions~\citep{kwiatkowski2019natural} or \msmarco~\citep{nguyen2016msmarco} to train the retriever directly. 
The key insight of \name is to amplify the power of few-shot examples by creating task-specific prompting, 
which in turn enables generating a large set of synthetic queries for training
retrievers suited for the task.
To ensure the generated data quality, we develop a filtering technique that ensures round-trip consistency {\em using generated data only}~\cref{sec:rt_filtering}.
Our filter is tailored to retrieval, which removes ambiguous, generic, and low-quality questions, and significantly improves retrieval performance.

 While \name is not the first application of LLM for retrieval, prior attempts of using LLMs often come with higher serving cost. \cite{Neelakantan2022TextAC} proposes to use the GPT-3~\citep{gpt3} embeddings in dual encoder models. However, the embedding size is 12k and hence makes the search index footprint and inference cost high. \cite{sachan2022improving} and~\cite{Bonifacio2022InParsDA} have applied prompting and LLMs for reranking, while leaving the retriever untouched. 
With \name, we show that LLMs can be used to generate efficient end-to-end retriever with high accuracy.
The contributions of the paper are as follows:
\vspace{-.05in}
\begin{itemize}
    \item We analyze the previously overlooked differences  across retrieval tasks in their search intents and query distributions, and propose a Few-Shot Retrieval setting for the BEIR dataset. Our prompt and fewshot examples will be released to facilitate future research.
    \item We propose \name, a simple recipe for few-shot retrieval by prompting with a LLM to generate synthetic task-specific training data. For the first time, end-to-end retrievers solely based on a few supervised examples can be strong and efficient to serve with \name.
    \item 
    Our experimental results show that, surprisingly,  \name with two-to-eight examples produced
    significantly better retrievers compared to recent models trained on \msmarco~ or NQ that have over 500K human annotated examples (Figure~\ref{fig:fewshot}(c)). \name outperforms ColBERT v2 and SPLADE v2 on 11 retrieval tasks we tested, while reranking boosts results by another 5 points on standard retrieval evaluation metric.
\end{itemize}

\section{Few-shot Retrieval Task}
\label{sec:task}

In this section, we first introduce the definition of a retrieval task and the differences among different retrieval tasks. We then propose a new Few-Shot Retrieval setting for the BEIR benchmark.

\subsection{Retrieval Task}

Given a large corpus, a retrieval model is responsible to find documents that are most relevant to a provided query $q$ according to a pre-defined relevancy. 
Formally, we define a retrieval task as: 
\begin{equation*}
    T = \{\mathcal{D}, \mathcal{Q}, \mathcal{I}\},
\end{equation*}
where $\mathcal{D}=\{d_1, d_2, ..., d_n\}$ is a large corpus of documents for retrieval,  $\mathcal{Q}$ is a query distribution, and $\mathcal{I}$ is the underlying search intent for the task. Depending on the task, $\mathcal{D}$ can be any document collection, such as web or Wikipedia. $Q$ also varies across tasks, e.g., short keyword search queries, questions, arguments, etc. If $\mathcal{I}(q,d)=1$, it means
 search intent of $q$ has been satisfied by the document $d$. For example, in QA tasks such as Natural Questions (\nq{}) the search intent is to find passages that provide the answer to the question, meaning $\mathcal{I}_\text{NQ}(q,d) = 1$ if $d$ answers $q$. Importantly, for the same pair of $(q,d)$, their relevance can be completely different under different search intents. For example, some argument retrieval tasks look for supporting arguments, while other tasks need to retrieve {\em counter} arguments.

In this work, we target the scenario where a target retrieval corpus $\mathcal{D_T}$ is available, but the amount of annotated query-document pairs for the new task is limited. Most prior of research efforts were put into adapting retrievers to new corpus $\mathcal{D_T}$,  but the divergence in queries $\mathcal{Q_T} $ and intents $\mathcal{I_T}$ remains under-explored. Next, we explore how search intent can be expressed with a short description and very few number of examples.

\subsection{Few-shot BEIR Setting} 

In this paper, we argue that it is important to let retrievers be aware of task-specific query distribution  and  search intent, as opposed to merely focusing on the domain adaptation of $\mathcal{D} $.  Prior belief is that it is expensive to collect enough in-distribution queries and relevance labels to train a neural retriever,  but intuitively, a person can understand a retrieval task by reading a short instruction and going over a few examples. In this work, we ask if  
 \emph{a few (8 or fewer)} examples are sufficient for the machines to learn a task-specific retriever. 
To facilitate our study and future research of few-shot retrieval, we define a new few-shot retrieval evaluation setting built upon the BEIR heterogeneous retrieval benchmark~\citep{thakur2021beir}. 

BEIR has 18 information retrieval datasets 
across 9 domains, including \textit{Bio-Medical}, \textit{Finance}, \textit{News}, \textit{Twitter}, \textit{Wikipedia}, \textit{StackExchange}, \textit{Quora}, \textit{Scientific}, and \textit{Misc}. These datasets also cover a diverse range of search intents: QA retrieval (question-to-document), duplicate question discovery (question-to-question), fact checking (claim-to-document), etc. Following~\cite{Santhanam2022ColBERTv2EA} and~\cite{spladev2}, we narrow our focus to the %
publicly-available datasets in BEIR.
The original BEIR evaluation used a zero-shot set up, where no queries or relevant query-document pairs from the evaluation datasets can be used to train the retrievers.  %

We extend BEIR to the few-shot setting by \emph{randomly} taking a few (2 to 8) in-domain relevant query-document examples as the task-specific supervision.
The examples are sampled from the development set when it is available.  For the BEIR tasks which only have a test set, we use samples from the test data as few-shot examples. To make the evaluation fair,  when evaluating few-shot retriever models, these test-set examples should be treated as `failed to retrieve` even if the model successfully retrieves them.  The prompt and few-shot examples will be released to the public.

\section{\textsc{Promptagator}}
\label{sec:model}

To approach the goal of creating retrievers from few-shot examples, we propose \fullname (\name).  The key idea of \name is to transform the few examples into many more examples by prompting a LLM, instead of using them to train a retriever directly. 

 \name consists of three components: prompt-based query generation, consistency filtering, and retriever training.  During prompt-based query generation, a task-specific prompt will be combined with a large language model to produce queries for the target task using $\mathcal{D}_T$. Then a filtering step cleans the generated data based on round-trip consistency; surprisingly, we found a retriever trained only on the synthetic data can be used to filter the synthetic data. Finally, a retriever (in this paper, dual encoders) and a cross attention reranker will be trained based on the generated data. Figure~\ref{fig:overall} in Appendix shows the overall procedure. 

\subsection{Prompt-base Query Generation}
\label{sec:prompt_qgen}

\name constructs instruction prompt that consists task-specific query passage descriptions and $k$  annotated query-document examples from the target dataset.
Specifically, let$\{(q_i, d_i)\}^k$ be $k$ relevant query-document pairs from the target task $T$, where $q_i \sim \mathcal{Q}_T$, $d_i \in \mathcal{D}_T$, and $\mathcal{I}_T(q_i, d_i)=1$.  Following FLAN~\citep{flan}, we use instruction prompts with the following form:
\begin{equation*}
(e_{doc}(d_i), e_{query}(q_1), \ldots, e_{doc}(d_k), e_{query}(q_k), e_{doc}(d))
\end{equation*}
where $e_{doc}(d)$ and $e_{query}(q)$ are task-specific document, query descriptions respectively, and $d$ is a new document.  Take ArguAna for example, we set $e_{doc}(d)=$ ``$\nlp{Argument:} \{d\}$''  and $e_{query}=$ ``$\nlp{Counter Argument:} \{q\}$'' to inform the LLM to generate counter arguments.  The LLM will is expected to generate $e_{query}(\hat{q})$.
If the LLM does not generate query description correctly, we consider it a generation failure and drop the output; otherwise we accept $q$ and form a synthetic relevant example $(q, d)$.

Running the prompt on all documents from $\mathcal{D}_T$, we can create a large set of synthetic $(q, d)$ examples, amplifying the information from few examples into a large synthetic dataset whose  query distribution is similar to true task distribution $\mathcal{Q}_T$ and  query-document pairs convey the true search intent $\mathcal{I}_T$.  %

We use FLAN~\citep{flan} as the LLM for query generation in this work. FLAN is trained on a collection of tasks described via instructions and was shown to have good zero/few-shot performance on unseen tasks. We use the 137B FLAN checkpoint provided by the authors. During prompt engineering, we use at most 8 examples, and reduce the number if they exceed the input length limit of FLAN. we also manually truncate individual query and document in the examples if they are too long.  We randomly sample up to 1 million documents from each corpus and generate 8 questions per document using sampling decoding with temperature 0.7.
The set of templates can be found in Table~\ref{tab:promt_template} in the Appendix.

\subsection{Consistency Filtering using only generated data}
\label{sec:rt_filtering}

The filtering step improves the quality of generated queries by ensuring the round-trip consistency~\citep{alberti-etal-2019-synthetic}: a query should be answered by the passage from which the query was generated. In our retrieval case, the query should retrieve its source passage.
Consistency filtering~\citep{alberti-etal-2019-synthetic, paq} has been shown crucial for synthetic question generation on QA tasks.  However,
these techniques typically rely on an \emph{external} question-answering model as the filter, trained on existing supervised QA data. Since we want to address different search intents, using a single external filtering model does not work for us. %

Surprisingly, we find out that consistency filtering based on the {\em generated data alone}
can work well over the different search intents observed in BEIR.
We first use the {\em generated query and document pairs} to train an initial retriever.
Given a synthetic query-document pair ($q$, $d$), we use the initial retriever to predict the most relevant passages for $q$.
We keep $q$ only when $d$ occurs among the Top-$K$ passages returned by the retriever. This may seem unintuitive because the filtering model (the initial retriever) is trained on the same noisy synthetic data that it will filter.
We show this filter substantially reduces the number of synthetic queries and significantly improves retrieval performance.

\subsection{Few-shot \name Retriever}

Our synthetically generated data allows training task-specific neutral retrievers for tasks where supervised in-domain fine-tuning is challenging due to data scarcity. In this work, we use the standard dual-encoder retrieval architecture and we propose a simple pretrain/fine-tune recipe.

Following prior work~\citep{DBLP:journals/corr/abs-2112-07899}, we initialize the dual encoder using the Transformer encoder from a  T5~\citep{2020t5} checkpoint. 
We then pretrain our retriever on C4 with the independent cropping task from Contriever~\citep{izacard2021contriever}, where we treat two random crops from the same document as positive retrieval pairs and train with a cross-entropy loss over in-batch random negatives. 
Next, we fine-tune the dual-encoder on the query-document pairs generated from our prompt-base QGen, again with cross-entropy loss over in-batch random negatives. After training for a set number of epochs, we apply round-trip filtering on our synthetic data as described in \cref{sec:rt_filtering} using this initial dual encoder, and continue to fine-tune the dual encoder on the filtered data.

We also propose \namep, a reranker trained on the same synthetic data generated from our prompt-base 
QGen, which refines the retrieved candidates using a slower but more accurate cross-attention model. We train the reranker using a cross-entropy loss with 31 sampled negatives from top 200 passages retrieved by the \name retriever, which approximates the inference time distribution (reranking top 200 from the retriever).

\subsection{Zero-shot \name retriever}
The prompt-based query generation can also run in a zero-shot  manner, where we universally apply the following prompt irrespective of the target task: 
\nlp{f'\{d\} Read the passage and generate a query.'}.    
Here $d$ denotes the document text.  We train retrievers and rerankers on the zero-shot prompt generated data, leading to zero-shot \name and zero-shot \namep.

\subsection{Discussion}

Table \ref{tab:comparisons} compares the \name recipe to some recently proposed approaches. Our dual encoder does not rely on \emph{hard negative mining} or \emph{distillation}; it uses a standard dual encoder model without adding the token-level matching inductive biases that ColBERT and SPLADE have. Our reranker also uses a 110M model instead of larger models. We aim to use this simplified recipe to highlight the power of few-shot data, as we will shown in~\cref{sec:exp}.  Comparing \name to these approaches, the ability to use a prompt and few-shot examples with a LLM makes \name be able to generate efficient models with high accuracy.
While other LLM approaches such as  InPars~\citep{Bonifacio2022InParsDA} and UPR~\citep{sachan2022improving} have focused on reranking, \name focuses on retrieval.

\section{Experiments}

\begin{table}
\small
\centering
\resizebox{.9\linewidth}{!}{%
\begin{tabular}{c|cccccccc}
\toprule
           & \multicolumn{1}{c}{\begin{tabular}[c]{@{}c@{}}Retrieval\\ Supervision\end{tabular}} & \multicolumn{1}{c}{\begin{tabular}[c]{@{}c@{}}Cross-Attn \\ Distillation\end{tabular}} & \multicolumn{1}{c}{Retriever} & \multicolumn{1}{c}{\begin{tabular}[c]{@{}c@{}}Token-level\\ Retrieval\end{tabular}} &  \multicolumn{1}{c}{\begin{tabular}[c]{@{}c@{}}Serving Model\\ Size\end{tabular}}  & \multicolumn{1}{c}{\begin{tabular}[c]{@{}c@{}}\# Reranking\\ Doc.\end{tabular}}   & \multicolumn{1}{c}{\begin{tabular}[c]{@{}c@{}} QGen\\ Model\end{tabular}} \\
\midrule
Contriever    &  NA      &          &   self   &           & 110M       &   0    \\

GTR-XXL    &  \msmarco (500K)       &           &   self        &            &    6B     &   0    \\
Splade v2  &  \msmarco  (500K)      & \cmark    &   self        & \cmark     &   110M   &   0    \\
ColBERT v2 &  \msmarco  (500K)      & \cmark    &   self        & \cmark    &    110M   &   0    \\
GenQ       &  \msmarco (500K)       & \cmark    &   self        &           &   110M   &   0    & T5 (\msmarco)\\ 
GPL        &  \msmarco (500K)       & \cmark    &   self        &           &   110M   &   0    & T5 (\msmarco)\\ 
MonoT5     &  \msmarco (500K)       &           &   BM25        & \cmark    &    3B     &  1000   \\ 
InPars     &  Few (3)                   &           &   BM25        & \cmark    &    3B     &  1000     & GPT-3\\ 
UPR        &  NA                    &           &   Contriever  &        &   110M+3B     &  1000 & T0$^*$ \\ \midrule

\name   & Few  (0-8)           &     &  self    &          &  110M     &   0              & FLAN\\
\namep   & Few (0-8)         &     &  \name    &          &   110M+110M     &  200              & FLAN\\
\bottomrule
\end{tabular}
}
\caption{
\label{tab:comparisons} Comparison of settings, resources and model size for  different frameworks. Our models are just a 110M-size dual encoder \name and a 110M-size reranker \namep, as good quality generated data allows simple models/pipeline to achieve strong performance.  See text for more details for UPR's QGen model\protect\footnotemark.
}
\end{table}

\footnotetext{UPR uses T0 query generation for reranking, instead of for synthetic data augmentation that other QGen approaches do.}
We report quantitative evaluation of \name by measuring its retrieval performance on the BEIR benchmark. We then dive deeper into the results through ablation studies and qualitative analysis. 

\subsection{Implementation}
The original FLAN training set overlapped with 2 datasets in the BEIR benchmark: NQ\footnote{FLAN is only trained on question-to-answer tasks and never observes the question-passage supervision needed for retrieval training. Additionally, FLAN has not been fine-tuned on query generation tasks on QA datasets. } and Quora\footnote{We study the impact of NQ and Quora on FLAN query generation in \cref{subsec:ablation}}. Most of existing systems use all of the supervised data from \msmarco~ in their system. Therefore we exclude \msmarco, NQ and Quora from our main evaluations.  We report nDCG@10, the standard retrieval evaluation metric on BEIR.

For \name's prompt-based query generation, we sample  questions from the LLM with a temperature of 0.7.  
For round-trip filtering, we use \msmarco~as validation set and tune $K$. 
We find setting $K$ to 1 leads to the best results and thus use 1 for all BEIR datasets, i.e. we keep a $(q, d)$ pair only when $d$ is ranked in the top 1 place by the initial dual encoder.
 
We implement  \name's dual encoders following GTR~\citep{DBLP:journals/corr/abs-2112-07899};  in particular, we use a shared Transformer encoder initialized from T5, take the mean pooling of the top encoder layer, and project it to a fixed 768-dimensional embedding. To ensure efficiency, we use the T5-base version 1.1 encoder architecture consisting of 110M parameters. For \namep reranking models, we use the standard Transformer cross attention encoder, also initialized with a 110M T5-base encoder checkpoint.  At inference time, we rerank the top 200 candidates retrieved from the \name dual encoder retriever.  

We mostly follow the hyper-parameters used in the~\cite{DBLP:journals/corr/abs-2112-07899}. The default batch size in this recipe is 6k; however, some of the corpora in BEIR contain only a few thousand documents, making multiple relevant documents appear in the same batch which interacts negatively with the in-batch softmax loss. We found it important to use appropriate batch sizes and training steps for those small datasets.
We split the datasets into three groups based on corpus size:
small datasets (<50k), middle datasets (50k-500k), large datasets (>500k). For dual encoder training, we use 128 batch size for small datasets and 6k for others. We finetune for 5k steps for large datasets and 1k for others. For ranking models, we use batch size of 64 for all datasets and finetune large datasets for 20k steps, 5k for others. %

\begin{table*}
\small
  \centering
  \resizebox{\linewidth}{!}{%
\begin{tabular}{cccccccccccc|c}
\toprule 
 & arg & touché & covid &nfc & hotpot & dbp & climate & fever & scifact & scidocs & fiqa & AVG. \\ \midrule  

\multicolumn{13}{c}{Retriever} \\
\midrule
\multicolumn{13}{l}{{\textit{Unsupervised}}}  \\

BM25                 & 31.5                     & \textbf{36.7}                       & 65.6                      & 32.5                    & 60.3                       & 31.3                     & 21.3                        & 75.3                      & 66.5                        & 15.8                        & 23.6                     & 41.8                     \\
Contriever           & 37.9                     & 19.3                       & 27.4                      & 31.7                    & 48.1                       & 29.2                     & 15.5                        & 68.2                      & 64.9                        & 14.9                        & 24.5                     & 34.7                     \\  
\multicolumn{13}{l}{{\textit{Supervised [MS MARCO]}}}  \\
GTR-XXL              & 54.0                     & 25.6                       & 50.1                      & 34.2                    & 59.9                       & 40.8                     & \textbf{26.7}                        & 74.0                      & 66.2                        & 16.1                        & 46.7                     & 44.9                     \\
SPLADE v2            & 47.9                     & 27.2                       & 71.0                      & 33.4                    & \textbf{68.4}                       &\underline{ 43.5}                     & 23.5                        & \textbf{78.6}                      & \underline{69.3}                        & 15.8                        & 33.6                     & \underline{46.6}                     \\
ColBERT v2            & 46.3                     & 26.3                       & 73.8                      & 33.8                    & \underline{66.7}                       & \textbf{44.6}                     & 17.6                        & \underline{78.5}                      & \textbf{69.3}                        & 15.4                        & 35.6                     & 46.2    \\                 
GenQ                & 49.3                     & 18.2                       & 61.9                      & 31.9                    & 53.4                       & 32.8                     & 17.5                        & 66.9                      & 64.4                        & 14.3                        & 30.8                     & 40.1                     \\
GPL                  & \underline{55.7}                     & 25.5                       & 70.0                      & \textbf{34.5}                    & 58.2                       & 38.4                     & 23.5                        & 75.9                      & 67.4                        & \underline{16.9}                        & 34.4                     & 45.5                     \\ \cmidrule(lr){1-1} 
\multicolumn{13}{l}{\textbf{\name} (110M)  }                                                                                                                                                                                                                                                                                                                                      \\
Zero-shot           & 53.8                    & 26.6                      & \underline{72.7}                     & \textbf{33.4}                   & 60.4                      & 36.4                    & 21.4                       & 76.2                     & 62.3                       & 16.3                       & \underline{40.4}                    & 45.5\\
Few-shot         & \textbf{59.4}                    & \underline{34.5}                      & \textbf{75.6}                     & 33.4                   & 61.4                      & 38.0                    & 16.8 (24.0$^*$)                    & 77.0                     & 65.0                       & \textbf{18.4}                       & \textbf{46.2}                    & \textbf{47.8}                    \\ \midrule

\multicolumn{13}{c}{Retriever $+$ Reranker} \\
\midrule
\multicolumn{13}{l}{{\textit{Unsupervised}}}  \\
UPR (3B)       & 50.3                     & 21.3                       & 60.4                      & 33.3                    & 72.2                       & 33.8                     & 9.5                         & 57.3                      & 69.6                        & 17.3                        & 45.0                     & 42.7                      \\
InPars (3B) & -- & -- & \underline{78.4} & -- & -- & -- &-- &-- &-- &-- &-- &-- \\ 
\multicolumn{13}{l}{{\textit{Supervised [MS MARCO]}}}  \\
monoT5 (220M)         & 13.2                    & 27.7                      & 77.8                           &    35.7                     & 69.5                           & 41.9                         &   \underline{24.5} & 80.2                          &  \underline{73.6}                         &   16.5                          &    41.4                                                  & 45.6                    \\
monoT5 (3B)           & 28.8                    & 20.0                      & \textbf{79.5}                     & \textbf{38.4}                   & \textbf{75.9}                      & \textbf{47.8}                    & \textbf{28.0}                       & \underline{85.0}                     & \textbf{77.7}                       & \underline{19.7}                       & \textbf{51.4}                    & \underline{51.1}                    \\ \cmidrule(lr){1-1} 
\multicolumn{13}{l}{\textbf{\namep} (110M + 110M)}                                                                                                                                                                                                                                                                                                                            \\
Zero-shot            & \underline{52.1}                    & \underline{27.8}                      & 76.0                     & 36.0                   & 71.2                      & 41.3                    & {22.6}                       & 83.8                     & 73.2                       & 19.1                       & 45.9                    & 49.9                    \\
Few-shot            & \textbf{63.0}                    & \textbf{38.1}                      & 76.2                     & \underline{37.0}                   & \underline{73.6}                      & \underline{43.4}                    & 20.3 (24.1$^*$)                      & \textbf{86.6}                     & 73.1                       & \textbf{20.1}                       & \underline{49.4}                    & \textbf{52.8}                   \\ \bottomrule
\end{tabular}
}
\caption{ %
{\bf Main Results.} nDCG@10 on BEIR.
{\bf Retriever Comparisons (Upper Half):}
Among the various kind of retrievers, both zero-shot and few-shot \name produce strong results .
Note that ColBERT v2 and SPLADE v2 allows token-level interactions, but \name DE models do not.
{\bf Retriever+Reranker Comparisons (Lower Half):}
In the scenario where speed is not a concern, reranker is often used. We train \namep use the {\em same} generated data and get significant improvement.  See text for more details for Climate-Fever.\protect\footnotemark}
 \label{tab:main_results}
\end{table*} 

\footnotetext{\climatefever's relevant $(q,d)$ pairs in BEIR are not well-defined \cref{subsec:ablation},  so we also tried running query-generation with \fever's few-shot prompt on \climatefever. We report the results with \fever{} prompt in $()$, but they are not used for computing the average. }

\subsection{Main Results}

Table~\ref{tab:main_results} shows the experimental results. 
We first notice that zero-shot \name already serves as a strong baseline, comparing favorably to other retrieval baselines trained on $\mathcal{O}$(100K) examples from \msmarco{}. Nonetheless, few-shot \name markedly improves upon zero-shot \name, increasing the averaged nDCG@10 by over 2 points, which highlights
the impact of adapting the LLM to the target task.
Few-shot \name, being relatively simple in training steps and model architecture,  outperforms strong  baselines such as GenQ \citep{thakur2021beir} and GPL \citep{wang-etal-2022-gpl} which also use query generation to augment training data, as well as ColBERT v2 \citep{Santhanam2022ColBERTv2EA} and SPLADE v2 \citep{spladev2} which rely on token level interaction architectures and distillation recipes. %

Our reranker \namep further boosts performance with another 5 points gain on nDCG@10. It significantly outperforms UPR~\citep{sachan2022improving} whose reranker uses T0~\citep{sanh2021tzero}, an instruction tuned LLM similar to FLAN.  It also outperforms monoT5-3B~\citep{Nogueira20monot5}, which  achieved previous state-of-the-art reranking performance on BEIR in a recent study by~\cite{rosa2022no}. Note most of these reranker approach uses a 3B model for its better generalization ability than smaller models, while \namep uses a standard 110M rereanker.

Comparing few-shot \name to baselines,  the biggest improvement
is on \touche{} (touché), followed by \arguana{} (arg) .  \touche{}'s goal is to retrieve documents for  a \emph{controversial} topic, e.g., \textit{``should felons who have completed their sentence be allowed to vote?''}. \arguana{}'s goal is to find the \emph{counter-arguments} that oppose the input argument, and the input arguments are often several-sentence long.  Both tasks are  extremely different from traditional QA retrieval data that other models use, which are dominated by
factoid questions. On the other hand, few-shot \name can successfully adapt to this task with a few examples.

\subsection{Ablation Study}
\label{subsec:ablation}

Next, we study our results in greater detail and analyze factors contributing to performance.

\begin{figure}
\centering
\includegraphics[width=.95\linewidth]{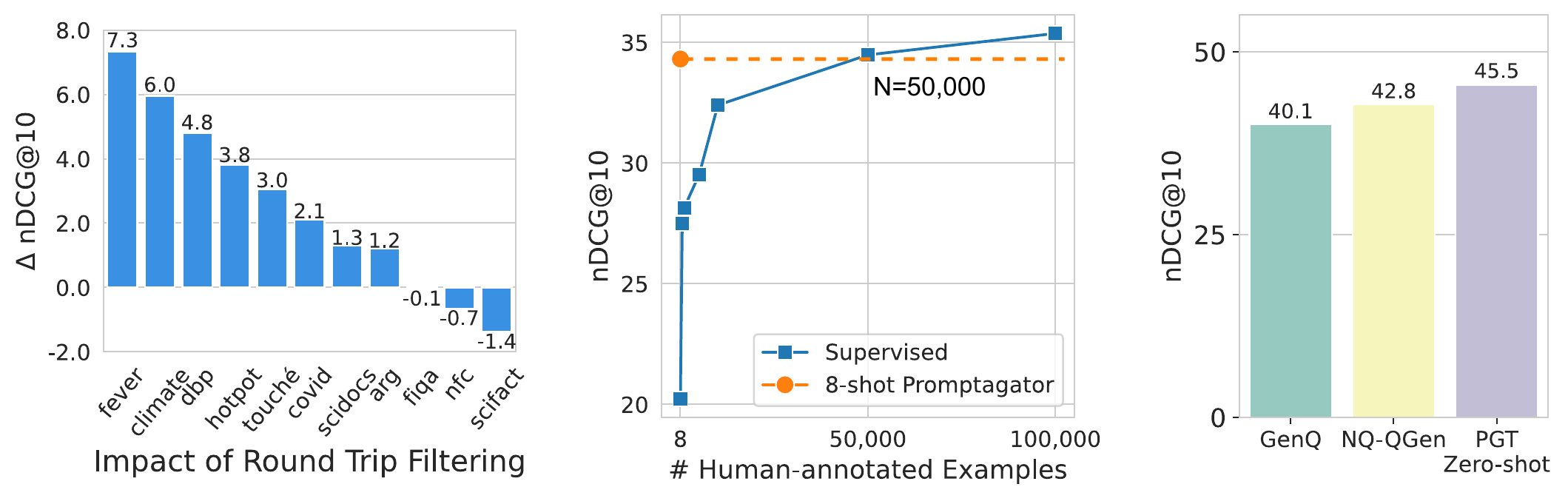}
\caption{{\bf Left (a): Consistency filter}. Delta in nDCG@10 between few-shot \name with and without round-trip filter. 
{\bf Middle (b)}: Comparing the effect of the generated data versus the number of supervised data on \msmarco{}. The LLM can amplify the power of the few examples, making 8 examples to catch up with 50k labeled examples, when simple dual encoders are used.  {\bf Right (c): } Ablation on query generation model.
GenQ is a prior query generation system from~\cite{thakur2021beir}, while NQ-QGen is our in-house T5 query generation model trained on NQ. Other than the generated data, NQ-QGen and \name uses the same hyper parameters.}
\label{fig:left_ex_right_filtering}
\end{figure}

\textbf{Impact of consistency filtering.}
In Figure~\ref{fig:left_ex_right_filtering}(a), we show quality difference between few-shot \name with and without round-trip filtering.
We can see that filtering improves performance on 8 out of 11 datasets, and leads to 2.5 points improvement on average.
These results demonstrate the effectiveness of our fitlering strategy.
There are nonetheless datasets where filtering hurts model quality such as NFCorpus and SciFact. 
Note these are the smallest datasets in terms of generated queries.
We conjecture further tuning dual encoder models on the filtered data results in overfitting.

Manually examining the query document pairs removed by the filter, we find the majority cases are either the query being too general which matches many documents, or the query contains hallucination irrelevant to the document.     
There are also cases where high quality query document pairs were incorrectly removed since the initial dual encoder model ranks other documents higher.
We suspect designing query-specific K values would help retain such query document pairs and further improve model performance.  
We leave this to future exploration.   

\textbf{Can generated queries replace human annotated queries?}
In Figure~\ref{fig:left_ex_right_filtering}(b), we compare the dual encoders trained on examples generated from the 8-shot \name vs the dual encoders trained on supervised data. Note that we did not add other components to make the comparison simple. We choose \msmarco{} as there are enough labeled data for this task and neither FLAN nor our models are trained on \msmarco{} examples. The results showed that the eight examples plus LLM can replace a significant portion of the supervised examples. 

\textbf{How does \name compare to other query generation approaches?} 
Figure~\ref{fig:left_ex_right_filtering}(c) compares zero-shot \name to two query generation approaches: GenQ~\citep{thakur2021beir} uses a \msmarco{} trained T5 query generation model, and NQ-QGen is our in-house T5 QGen model fine-tuned on NQ.  The figure shows the advantages of zero-shot \name, outperforming both fine-tuned QGen models by large margins. NQ-QGen uses the same filtering, dual-encoder training, batch sizes and training steps as \name, providing an apple-to-apple comparison of the query generators. These results indicate that the main contributing factor to \name is the prompted LLM query generation, not the specific training recipe or hype-parameters.

\paragraph{Does Few-shot always improve over Zero-shot?}
As shown in Table~\ref{tab:main_results},  few-shot \name almost always outperforms zero-shot \name. The only exception is \climatefever{} (climate). After examining this dataset, we realized that in the original \climatefever{} dataset, a query-document pair can be annotated as either ``supports'', ``refutes'', or ``not enough info''. BEIR treats these three annotations all as relevant; however, a ``not enough info'' document may not be related to the query. Using such pairs in the few-shot prompts can hurt query generation.  Therefore, we tried switching to \fever{}'s few-shot prompt, as the two datasets share same corpus and similar search intents. With the better annotated examples,  few-shot \name indeed surpass zero-shot. This result provides some evidence that low quality few-shot examples negatively affect \name.  

\paragraph{Impact of FLAN versions}
\begin{table}[h]
\small
  \centering
  \resizebox{\linewidth}{!}{%
\begin{tabular}{lccccccccccc|c}
\toprule 
 & arg & touché & covid &nfc & hotpot & dbp & climate & fever & scifact & scidocs & fiqa & AVG. \\ \midrule  
\midrule
FLAN original     & 59.4 & 34.5 & 75.6 & 33.4 & 61.4 & 38.0 & (24.0*) & 77.0 & 65.0 & 18.4 & 46.2 & 48.5 \\
FLAN w/o NQ and Quora   & 58.8 & 33.3 & 70.2 & 33.7 & 61.7 & 34.4 & (23.5*) & 76.2 & 63.8 & 18.3 & 43.0 & 47.0\\ \bottomrule
\end{tabular} 
}
\caption{Impact of different FLAN version. We use Fever model for Climte Fever for this study. See~\cref{subsec:ablation} for more details.}%
\label{tab:nobeir_flan}
\end{table}
In the main experiments we have used the FLAN model described in \cite{flan}.  This model was trained on a collection of datasets including question-answer datasets; specifically, it includes Natural Questions (NQ) and Quora. FLAN was not trained on query-document pairs from NQ or Quora; however, in order to determine whether the inclusion of this data biased the results on the final retrieval evaluation, we designed an additional ablation experiment.  Following the recipe from \cite{flan} used to train the original FLAN models, we trained an additional LLM excluding both the NQ and Quora datasets.
Table~\ref{tab:nobeir_flan} shows the results for \name trained with and without NQ and Quora.   While the accuracy drops slightly, the overall performance still outperform prior retrievers.

\label{sec:exp}

\subsection{Qualitative Analysis}

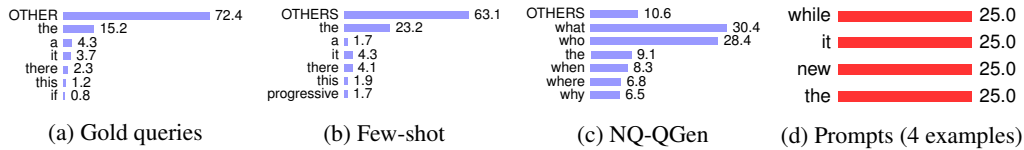
\begin{figure}
\centering
\begin{subfigure}[]{.24\textwidth}
\resizebox{\linewidth}{!}{
\begin{tikzpicture}[
  y=0.3cm,
  x=0.045cm,
]

\foreach [count=\i from 0] \p/\t in
                  {72.4/OTHER,
                   15.2/the,
                   4.3/a,
                   3.7/it,
                   2.3/there,
                   1.2/this,
                   0.8/if}
  {
   \node [anchor=base east,
          barlabels,
          name=i-\i] at (0,-\i) {\t};
   \fill [blue!40] (i-\i.base east) rectangle ++(\p,barheight)  ++(0,-barheight)
          node[barlabels, 
               black,
               anchor=base west] {\p};
  }

\end{tikzpicture}
}
\caption{Gold queries}
\label{fig:topwordsgold}
\end{subfigure}
\begin{subfigure}[]{.24\textwidth}
\resizebox{\linewidth}{!}{
\begin{tikzpicture}[
  y=0.3cm,
  x=0.045cm,
]

\foreach [count=\i from 0] \p/\t in
                  {63.1/OTHERS,
                   23.2/the,
                   1.7/a,
                   4.3/it,
                   4.1/there,
                   1.9/this,
                   1.7/progressive
                   }
  {
   \node [anchor=base east,
          barlabels,
          name=i-\i] at (0,-\i) {\t};
   \fill [blue!40] (i-\i.base east) rectangle ++(\p,barheight)  ++(0,-barheight)
          node[barlabels, 
               black,
               anchor=base west] {\p};
  }

\end{tikzpicture}
}
\caption{Few-shot}
\label{fig:topwordsfewshot}
\end{subfigure}
\begin{subfigure}[]{.24\textwidth}
\resizebox{\linewidth}{!}{
\begin{tikzpicture}[
  y=0.3cm,
  x=0.1cm,
]

\foreach [count=\i from 0] \p/\t in
                  {10.6/OTHERS,
                  30.4/what,
                   28.4/who,
                   9.1/the,
                   8.3/when,
                   6.8/where,
                   6.5/why}
  {
   \node [anchor=base east,
          barlabels,
          name=i-\i] at (0,-\i) {\t};
   \fill [blue!40] (i-\i.base east) rectangle ++(\p,barheight)  ++(0,-barheight)
          node[barlabels, 
               black,
               anchor=base west] {\p};
  }

\end{tikzpicture}
}
\caption{\nqqgen}
\label{fig:topwordsnq}
\end{subfigure}
\begin{subfigure}[]{.24\textwidth}
\resizebox{\linewidth}{!}{
\begin{tikzpicture}[
  y=0.45cm,
  x=0.09cm,
]

\foreach [count=\i from 0] \p/\t in
                  {25.0/while,
                  25.0/it,
                  25.0/new,
                  25.0/the}
  {
   \node [anchor=base east,
          barlabels,
          name=i-\i] at (0,-\i) {\t};
   \fill [red!20!black,fill=red!80!white] (i-\i.base east) rectangle ++(\p,barheight)  ++(0,-barheight)
          node[barlabels, 
               black,
               anchor=base west] {\p};
  }

\end{tikzpicture}
}
\caption{Prompts (4 examples)}
\label{fig:topwords_prompt}
\end{subfigure}
\caption{Top first word distribution on queries generated from different models in the ArguAna
dataset. 
{\bf Left (a)(b)(c)}: Compare gold queries (a) and generated queries (b)(c). Queries generated by few-shot models has closer distribution to the gold queries, while the \nqqgen{} queries are mostly questions. 
{\bf Right (d)}: The few shot FLAN can generate diverse queries even though there are only 4 examples in the prompt. Statistics of more datasets are available in the Appendix (Figure \ref{fig:topwords_appendix}).
\label{fig:topwords}
}
\end{figure}

In order to understand the advantages of few-shot \name , we analyze the distribution of the first words of the queries generated by different query generation methods for the ArguAna task in Figure~\ref{fig:topwords}. Note that the distribution of few-shot \name (Fig.~\ref{fig:topwordsfewshot}) is much closer to the real distribution (Fig.~\ref{fig:topwordsgold}) while the \nqqgen~ (Fig.~\ref{fig:topwordsnq}) mostly generated questions even when query of the tasks are arguments.
Examples are showcased in Table~\ref{tab:examples} in the Appendix. %

\section{Related Work}

\textbf{Neural retrieval models} The success of pre-trained large language models \citep{Devlin2019BERTPO, 2020t5, gpt3} has fostered a lush growth in the field of neural retrieval models.   
Neural retrieval models can be grouped into two categories, namely representation based models and interaction based models.

Representation based models \citep{palangi2016deep, GillickE2E, dense-passage-qa} encode a query and passage independently into a common dense space, and scores their relevance based on vector dot-product or cosine similarity. 
Recent research on representation based models has primarily focused on the following aspects:
developing better pre-training tasks \citep{lee-etal-2019-latent, Chang2020Pre-training, Colbert, izacard2021contriever, oguz-etal-2022-domain} or pre-training architectures \citep{gao-callan-2021-condenser, gao-callan-2022-cocondenser}, improving expressiveness using multi-vector representations \citep{luan-2021-sparse}, 
improving negative contrast \citep{rocketqa, ance, lu-etal-2021-multi}, and improving generalization across different domains \citep{thakur2021beir, ren2022dreval}. Different techniques have been explored to improve the generalization, such as using query generation for data augmentation \citep{ma-etal-2021-zero,wang-etal-2022-gpl}, using contrastive learning for better pre-training \citep{izacard2021contriever}, using knowledge distillation \citep{chen2021-salient, wang-etal-2022-gpl}
and scaling the model size \citep{DBLP:journals/corr/abs-2112-07899}. 

Although encoding the query and document into a single vector enables fast retrieval via approximate nearest neighbor search \citep{DBLP:journals/corr/JohnsonDJ17, wu2019efficient}, it also constrains the representational power of these models thus leading to sub-optimal predictions. Interaction based models on the other hand explicitly model the interaction between query and document terms \citep{Guo2016ADR,hui2017pacrr,xiong2017end,dai2018convolutional,DeepRelevenceRanking,nogueira2019passage}, and therefore make more informed decisions.  
These models are typically more expensive, and thus are used for reranking or rescoring.
Distilling interaction based models into representation based models has been shown effective in closing the gap between the two \citep{Hofsttter2020ImprovingEN,ren-etal-2021-pair,lin-etal-2021-batch,ren-etal-2021-rocketqav2, DBLP:conf/aistats/ReddiPMRYKVK21, zhang2022adversarial}.  
Another attempt to combine the best of both worlds is by postponing the interaction until the last layer of the model \citep{gao-etal-2021-coil, Colbert}, blurring the boundary between representation and interaction models. 

\paragraph{Few-shot Learning}The development of pre-trained large language models also popularize the few-shot learning paradigm, which utilizes a few examples as context for model inputs \citep{gpt3, wei-2022-emergent}. Two approaches are commonly used. One approach is to provide the LLM an instruction of the task in natural language with a few examples and do not update any parameter of LLM \citep{gpt3,Bonifacio2022InParsDA}. The other approach provides the LLM the instruction, a few examples and also performs model fine-tuning \citep{schick-schutze-2021-exploiting,schick-schutze-2021-just,schick-schutze-2021-shot,gao-etal-2021-making,logan-iv-etal-2022-cutting,Izacard-2022-fewshot}. Our work adopts the first approach. Usually 10-100 examples are used. For example, 32 examples are used in the few-shot setting in GPT3. 
In the context of retrieval, ~\cite{Bonifacio2022InParsDA} provides GPT3 three question-document pairs and uses it as the question generator for training interaction based models.

\paragraph{Prompt-based Query Generation}
The idea of using prompted LLMs for query generation has previously been proposed for improving retrieval reranking. In UPR~\citep{sachan2022improving}, they proposed to use prompted LLMs to rerank the passages directly. InPars~\citep{Bonifacio2022InParsDA} is probably the most closely related work to ours, where they proposed to use few-shot prompting with GPT-3 to generate synthetic data for training a T5-based reranker applied to a BM25 retriever. In this paper, we propose few-shot prompted LLMs and show that generated data can produce efficient and strong end-to-end retrievers. Moreover, we show the quality of generated data can be improved by task-specific prompts and consistency filtering.

\paragraph{Retrievers with late interactions} While dual encoder models are very efficient at retrieval due to the MIPS algorithms, their expressivity is limited due to the fact that their score is just a dot-product between the query vector and the document vector. 
ColBERT~\citep{Santhanam2022ColBERTv2EA, Colbert} and SPLADE~\citep{spladev2} are the models to increase the interactions between the query and document by allowing token-level interactions. Because these models are not just dot product between queries and documents, MIPS algorithms can not be used directly. Hence, these models usually have much higher serving cost compared to dual encoders.

\section{Conclusion and Discussions}
\label{sec:discussion}

In  this  paper,  we  have  presented \name,  a novel approach to few-shot retrieval. We  showed  that  it  is  possible  to  create task-specific, end-to-end retrievers with only a few annotated examples.  The few-shot examples, amplified by prompt-based LLM query generation, simplifies the complexity of training neural retrievers for a new tasks and leads to promising retrieval performance gains. It hopefully inspires future research to further push the limit of few-shot retrieval, towards generalizable retrieval systems that can seamlessly and efficiently adapt to many tasks.

While we demonstrate that query generation can be very effective, many questions remain for the roles of question generation and large language models. One of the key issue that needs thorough investigation is on the {generated data efficiency}. We have not yet explored exactly how many query-document pairs are needed for each task, or how to use these generated examples more efficiently. Another issue that is worthwhile understanding is the sensitivity of final retriever's performance with respect to the prompt. Finally, we would like to draw a connection from \name to distillation, as the final dual encoders definitely benefit a lot from the large language model. Analyzing the headroom and understanding how we can better transfer knowledge from LLMs to retrievers would be a critical topic for the future.

\section{Compute Usage and Environmental Impact}
We used the 137B large language model \flan{} for query generation. \flan{} is based on the same pretrained model as LaMDA \citep{lamda}. LaMDA was pre-trained on a large corpus consisting of 1.56T words, costing 451 MWh energy and 25.2 tCO2e carbon footprint. In \name{}, we generated 29.23M queries * 2 prompts = 58.46M queries, for a total of 610M words. 
As mentioned in \cref{sec:discussion}, \name{} can be viewed as distilling LLM to standard-sized dual encoders via prompt-based query generation. While the distillation process is computationally expensive, it significantly reduces cost for inference.

\section*{Acknowledgements}
We thank Kenton Lee, Tom Kwiatkowski, and Daniel Gillick for technical discussion and providing feedback on our manuscript. We thank Alex Salcianu for developing a bulk inference pipeline for large language models. 

\bibliography{cite}

\begin{thebibliography}{65}
\providecommand{\natexlab}[1]{#1}
\providecommand{\url}[1]{\texttt{#1}}
\expandafter\ifx\csname urlstyle\endcsname\relax
  \providecommand{\doi}[1]{doi: #1}\else
  \providecommand{\doi}{doi: \begingroup \urlstyle{rm}\Url}\fi

\bibitem[Alberti et~al.(2019)Alberti, Andor, Pitler, Devlin, and
  Collins]{alberti-etal-2019-synthetic}
Chris Alberti, Daniel Andor, Emily Pitler, Jacob Devlin, and Michael Collins.
\newblock Synthetic {QA} corpora generation with roundtrip consistency.
\newblock In \emph{Proceedings of the 57th Annual Meeting of the Association
  for Computational Linguistics}, pp.\  6168--6173, Florence, Italy, July 2019.
  Association for Computational Linguistics.
\newblock \doi{10.18653/v1/P19-1620}.
\newblock URL \url{https://aclanthology.org/P19-1620}.

\bibitem[Bonifacio et~al.(2022)Bonifacio, Abonizio, Fadaee, and
  Nogueira]{Bonifacio2022InParsDA}
Luiz Bonifacio, Hugo Abonizio, Marzieh Fadaee, and Rodrigo Nogueira.
\newblock Inpars: Unsupervised dataset generation for information retrieval.
\newblock In \emph{Proceedings of the 45th International ACM SIGIR Conference
  on Research and Development in Information Retrieval}, SIGIR '22, pp.\
  2387–2392, New York, NY, USA, 2022. Association for Computing Machinery.
\newblock ISBN 9781450387323.
\newblock \doi{10.1145/3477495.3531863}.
\newblock URL \url{https://doi.org/10.1145/3477495.3531863}.

\bibitem[Brown et~al.(2020)Brown, Mann, Ryder, Subbiah, Kaplan, Dhariwal,
  Neelakantan, Shyam, Sastry, Askell, Agarwal, Herbert-Voss, Krueger, Henighan,
  Child, Ramesh, Ziegler, Wu, Winter, Hesse, Chen, Sigler, Litwin, Gray, Chess,
  Clark, Berner, McCandlish, Radford, Sutskever, and Amodei]{gpt3}
Tom Brown, Benjamin Mann, Nick Ryder, Melanie Subbiah, Jared~D Kaplan, Prafulla
  Dhariwal, Arvind Neelakantan, Pranav Shyam, Girish Sastry, Amanda Askell,
  Sandhini Agarwal, Ariel Herbert-Voss, Gretchen Krueger, Tom Henighan, Rewon
  Child, Aditya Ramesh, Daniel Ziegler, Jeffrey Wu, Clemens Winter, Chris
  Hesse, Mark Chen, Eric Sigler, Mateusz Litwin, Scott Gray, Benjamin Chess,
  Jack Clark, Christopher Berner, Sam McCandlish, Alec Radford, Ilya Sutskever,
  and Dario Amodei.
\newblock Language models are few-shot learners.
\newblock In H.~Larochelle, M.~Ranzato, R.~Hadsell, M.F. Balcan, and H.~Lin
  (eds.), \emph{Advances in Neural Information Processing Systems}, volume~33,
  pp.\  1877--1901. Curran Associates, Inc., 2020.
\newblock URL
  \url{https://proceedings.neurips.cc/paper/2020/file/1457c0d6bfcb4967418bfb8ac142f64a-Paper.pdf}.

\bibitem[Chang et~al.(2020)Chang, Yu, Chang, Yang, and
  Kumar]{Chang2020Pre-training}
Wei-Cheng Chang, Felix~X. Yu, Yin-Wen Chang, Yiming Yang, and Sanjiv Kumar.
\newblock Pre-training tasks for embedding-based large-scale retrieval.
\newblock In \emph{International Conference on Learning Representations}, 2020.
\newblock URL \url{https://openreview.net/forum?id=rkg-mA4FDr}.

\bibitem[Chen et~al.(2021)Chen, Lakhotia, Oğuz, Gupta, Lewis, Peshterliev,
  Mehdad, Gupta, and Yih]{chen2021-salient}
Xilun Chen, Kushal Lakhotia, Barlas Oğuz, Anchit Gupta, Patrick Lewis, Stan
  Peshterliev, Yashar Mehdad, Sonal Gupta, and Wen-tau Yih.
\newblock Salient phrase aware dense retrieval: Can a dense retriever imitate a
  sparse one?
\newblock \emph{CoRR}, 2021.
\newblock URL \url{https://arxiv.org/abs/2110.06918}.

\bibitem[Dai et~al.(2018)Dai, Xiong, Callan, and Liu]{dai2018convolutional}
Zhuyun Dai, Chenyan Xiong, Jamie Callan, and Zhiyuan Liu.
\newblock Convolutional neural networks for soft-matching n-grams in ad-hoc
  search.
\newblock In \emph{Proceedings of the Eleventh ACM International Conference on
  Web Search and Data Mining}, WSDM '18, pp.\  126–134, New York, NY, USA,
  2018. Association for Computing Machinery.
\newblock URL \url{https://doi.org/10.1145/3159652.3159659}.

\bibitem[Devlin et~al.(2019)Devlin, Chang, Lee, and
  Toutanova]{Devlin2019BERTPO}
Jacob Devlin, Ming-Wei Chang, Kenton Lee, and Kristina Toutanova.
\newblock {BERT}: Pre-training of deep bidirectional transformers for language
  understanding.
\newblock In \emph{Proceedings of the 2019 Conference of the North {A}merican
  Chapter of the Association for Computational Linguistics: Human Language
  Technologies, Volume 1 (Long and Short Papers)}, pp.\  4171--4186,
  Minneapolis, Minnesota, June 2019. Association for Computational Linguistics.
\newblock URL \url{https://aclanthology.org/N19-1423}.

\bibitem[Formal et~al.(2021)Formal, Lassance, Piwowarski, and
  Clinchant]{spladev2}
Thibault Formal, Carlos Lassance, Benjamin Piwowarski, and St{\'{e}}phane
  Clinchant.
\newblock {SPLADE} v2: Sparse lexical and expansion model for information
  retrieval.
\newblock \emph{CoRR}, abs/2109.10086, 2021.
\newblock URL \url{https://arxiv.org/abs/2109.10086}.

\bibitem[Gao \& Callan(2021)Gao and Callan]{gao-callan-2021-condenser}
Luyu Gao and Jamie Callan.
\newblock Condenser: a pre-training architecture for dense retrieval.
\newblock In \emph{Proceedings of the 2021 Conference on Empirical Methods in
  Natural Language Processing}, pp.\  981--993, Online and Punta Cana,
  Dominican Republic, November 2021. Association for Computational Linguistics.
\newblock \doi{10.18653/v1/2021.emnlp-main.75}.
\newblock URL \url{https://aclanthology.org/2021.emnlp-main.75}.

\bibitem[Gao \& Callan(2022)Gao and Callan]{gao-callan-2022-cocondenser}
Luyu Gao and Jamie Callan.
\newblock Unsupervised corpus aware language model pre-training for dense
  passage retrieval.
\newblock In \emph{Proceedings of the 60th Annual Meeting of the Association
  for Computational Linguistics (Volume 1: Long Papers)}, pp.\  2843--2853,
  Dublin, Ireland, May 2022. Association for Computational Linguistics.
\newblock \doi{10.18653/v1/2022.acl-long.203}.
\newblock URL \url{https://aclanthology.org/2022.acl-long.203}.

\bibitem[Gao et~al.(2021{\natexlab{a}})Gao, Dai, and
  Callan]{gao-etal-2021-coil}
Luyu Gao, Zhuyun Dai, and Jamie Callan.
\newblock {COIL}: Revisit exact lexical match in information retrieval with
  contextualized inverted list.
\newblock In \emph{Proceedings of the 2021 Conference of the North American
  Chapter of the Association for Computational Linguistics: Human Language
  Technologies}, pp.\  3030--3042, Online, June 2021{\natexlab{a}}. Association
  for Computational Linguistics.
\newblock \doi{10.18653/v1/2021.naacl-main.241}.
\newblock URL \url{https://aclanthology.org/2021.naacl-main.241}.

\bibitem[Gao et~al.(2021{\natexlab{b}})Gao, Fisch, and
  Chen]{gao-etal-2021-making}
Tianyu Gao, Adam Fisch, and Danqi Chen.
\newblock Making pre-trained language models better few-shot learners.
\newblock In \emph{Proceedings of the 59th Annual Meeting of the Association
  for Computational Linguistics and the 11th International Joint Conference on
  Natural Language Processing (Volume 1: Long Papers)}, pp.\  3816--3830,
  Online, August 2021{\natexlab{b}}. Association for Computational Linguistics.
\newblock \doi{10.18653/v1/2021.acl-long.295}.
\newblock URL \url{https://aclanthology.org/2021.acl-long.295}.

\bibitem[Gillick et~al.(2018)Gillick, Presta, and Tomar]{GillickE2E}
Daniel Gillick, Alessandro Presta, and Gaurav~Singh Tomar.
\newblock End-to-end retrieval in continuous space.
\newblock \emph{CoRR}, abs/1811.08008, 2018.
\newblock URL \url{https://arxiv.org/abs/1811.08008}.

\bibitem[Guo et~al.(2016)Guo, Fan, Ai, and Croft]{Guo2016ADR}
Jiafeng Guo, Yixing Fan, Qingyao Ai, and W.~Bruce Croft.
\newblock A deep relevance matching model for ad-hoc retrieval.
\newblock In \emph{Proceedings of the 25th ACM International on Conference on
  Information and Knowledge Management}, CIKM '16, pp.\  55–64, New York, NY,
  USA, 2016.
\newblock URL \url{https://doi.org/10.1145/2983323.2983769}.

\bibitem[Hasibi et~al.(2017)Hasibi, Nikolaev, Xiong, Balog, Bratsberg, Kotov,
  and Callan]{hasibi2017dbpedia}
Faegheh Hasibi, Fedor Nikolaev, Chenyan Xiong, Krisztian Balog, Svein~Erik
  Bratsberg, Alexander Kotov, and Jamie Callan.
\newblock Dbpedia-entity v2: A test collection for entity search.
\newblock In \emph{Proceedings of the 40th International ACM SIGIR Conference
  on Research and Development in Information Retrieval}, SIGIR '17, pp.\
  1265–1268, New York, NY, USA, 2017.
\newblock ISBN 9781450350228.
\newblock URL \url{https://doi.org/10.1145/3077136.3080751}.

\bibitem[Hofst{\"a}tter et~al.(2020)Hofst{\"a}tter, Althammer, Schr{\"o}der,
  Sertkan, and Hanbury]{Hofsttter2020ImprovingEN}
Sebastian Hofst{\"a}tter, Sophia Althammer, Michael Schr{\"o}der, Mete Sertkan,
  and Allan Hanbury.
\newblock Improving efficient neural ranking models with cross-architecture
  knowledge distillation.
\newblock \emph{ArXiv}, abs/2010.02666, 2020.
\newblock URL \url{https://arxiv.org/abs/2010.02666}.

\bibitem[Hui et~al.(2017)Hui, Yates, Berberich, and de~Melo]{hui2017pacrr}
Kai Hui, Andrew Yates, Klaus Berberich, and Gerard de~Melo.
\newblock {PACRR}: A position-aware neural {IR} model for relevance matching.
\newblock In \emph{Proceedings of the 2017 Conference on Empirical Methods in
  Natural Language Processing}, pp.\  1049--1058, Copenhagen, Denmark,
  September 2017. Association for Computational Linguistics.
\newblock \doi{10.18653/v1/D17-1110}.
\newblock URL \url{https://www.aclweb.org/anthology/D17-1110}.

\bibitem[Izacard et~al.(2022{\natexlab{a}})Izacard, Caron, Hosseini, Riedel,
  Bojanowski, Joulin, and Grave]{izacard2021contriever}
Gautier Izacard, Mathilde Caron, Lucas Hosseini, Sebastian Riedel, Piotr
  Bojanowski, Armand Joulin, and Edouard Grave.
\newblock Unsupervised dense information retrieval with contrastive learning.
\newblock \emph{Transactions on Machine Learning Research}, 2022{\natexlab{a}}.
\newblock URL \url{https://openreview.net/forum?id=jKN1pXi7b0}.

\bibitem[Izacard et~al.(2022{\natexlab{b}})Izacard, Lewis, Lomeli, Hosseini,
  Petroni, Schick, Dwivedi-Yu, Joulin, Riedel, and Grave]{Izacard-2022-fewshot}
Gautier Izacard, Patrick Lewis, Maria Lomeli, Lucas Hosseini, Fabio Petroni,
  Timo Schick, Jane Dwivedi-Yu, Armand Joulin, Sebastian Riedel, and Edouard
  Grave.
\newblock Few-shot learning with retrieval augmented language models,
  2022{\natexlab{b}}.
\newblock URL \url{https://arxiv.org/abs/2208.03299}.

\bibitem[Johnson et~al.(2021)Johnson, Douze, and
  Jégou]{DBLP:journals/corr/JohnsonDJ17}
Jeff Johnson, Matthijs Douze, and Hervé Jégou.
\newblock Billion-scale similarity search with gpus.
\newblock \emph{IEEE Transactions on Big Data}, 7\penalty0 (3):\penalty0
  535--547, 2021.
\newblock \doi{10.1109/TBDATA.2019.2921572}.
\newblock URL \url{https://doi.org/10.1109/TBDATA.2019.2921572}.

\bibitem[Karpukhin et~al.(2020)Karpukhin, Oguz, Min, Lewis, Wu, Edunov, Chen,
  and Yih]{dense-passage-qa}
Vladimir Karpukhin, Barlas Oguz, Sewon Min, Patrick Lewis, Ledell Wu, Sergey
  Edunov, Danqi Chen, and Wen-tau Yih.
\newblock Dense passage retrieval for open-domain question answering.
\newblock In \emph{Proceedings of the 2020 Conference on Empirical Methods in
  Natural Language Processing (EMNLP)}, pp.\  6769--6781, Online, November
  2020. Association for Computational Linguistics.
\newblock \doi{10.18653/v1/2020.emnlp-main.550}.
\newblock URL \url{https://aclanthology.org/2020.emnlp-main.550}.

\bibitem[Khattab \& Zaharia(2020)Khattab and Zaharia]{Colbert}
Omar Khattab and Matei Zaharia.
\newblock Col{BERT}: Efficient and effective passage search via contextualized
  late interaction over {BERT}.
\newblock In Jimmy Huang, Yi~Chang, Xueqi Cheng, Jaap Kamps, Vanessa Murdock,
  Ji{-}Rong Wen, and Yiqun Liu (eds.), \emph{Proceedings of the 43rd
  International {ACM} {SIGIR} conference on research and development in
  Information Retrieval, {SIGIR} 2020, Virtual Event, China, July 25-30, 2020},
  pp.\  39--48. {ACM}, 2020.
\newblock \doi{10.1145/3397271.3401075}.
\newblock URL \url{https://doi.org/10.1145/3397271.3401075}.

\bibitem[Kwiatkowski et~al.(2019)Kwiatkowski, Palomaki, Redfield, Collins,
  Parikh, Alberti, Epstein, Polosukhin, Devlin, Lee, Toutanova, Jones, Kelcey,
  Chang, Dai, Uszkoreit, Le, and Petrov]{kwiatkowski2019natural}
Tom Kwiatkowski, Jennimaria Palomaki, Olivia Redfield, Michael Collins, Ankur
  Parikh, Chris Alberti, Danielle Epstein, Illia Polosukhin, Jacob Devlin,
  Kenton Lee, Kristina Toutanova, Llion Jones, Matthew Kelcey, Ming-Wei Chang,
  Andrew~M. Dai, Jakob Uszkoreit, Quoc Le, and Slav Petrov.
\newblock Natural questions: A benchmark for question answering research.
\newblock \emph{Transactions of the Association for Computational Linguistics},
  7:\penalty0 452--466, March 2019.
\newblock \doi{10.1162/tacl_a_00276}.
\newblock URL \url{https://aclanthology.org/Q19-1026}.

\bibitem[Lee et~al.(2019)Lee, Chang, and Toutanova]{lee-etal-2019-latent}
Kenton Lee, Ming-Wei Chang, and Kristina Toutanova.
\newblock Latent retrieval for weakly supervised open domain question
  answering.
\newblock In \emph{Proceedings of the 57th Annual Meeting of the Association
  for Computational Linguistics}, pp.\  6086--6096, Florence, Italy, July 2019.
  Association for Computational Linguistics.
\newblock \doi{10.18653/v1/P19-1612}.
\newblock URL \url{https://aclanthology.org/P19-1612}.

\bibitem[Lewis et~al.(2021)Lewis, Wu, Liu, Minervini, K{\"u}ttler, Piktus,
  Stenetorp, and Riedel]{paq}
Patrick Lewis, Yuxiang Wu, Linqing Liu, Pasquale Minervini, Heinrich
  K{\"u}ttler, Aleksandra Piktus, Pontus Stenetorp, and Sebastian Riedel.
\newblock {PAQ}: 65 million probably-asked questions and what you can do with
  them.
\newblock \emph{Transactions of the Association for Computational Linguistics},
  9:\penalty0 1098--1115, 2021.
\newblock \doi{10.1162/tacl_a_00415}.
\newblock URL \url{https://aclanthology.org/2021.tacl-1.65}.

\bibitem[Lin et~al.(2021)Lin, Yang, and Lin]{lin-etal-2021-batch}
Sheng-Chieh Lin, Jheng-Hong Yang, and Jimmy Lin.
\newblock In-batch negatives for knowledge distillation with tightly-coupled
  teachers for dense retrieval.
\newblock In \emph{Proceedings of the 6th Workshop on Representation Learning
  for NLP (RepL4NLP-2021)}, pp.\  163--173, Online, August 2021. Association
  for Computational Linguistics.
\newblock \doi{10.18653/v1/2021.repl4nlp-1.17}.
\newblock URL \url{https://aclanthology.org/2021.repl4nlp-1.17}.

\bibitem[Logan~IV et~al.(2022)Logan~IV, Balazevic, Wallace, Petroni, Singh, and
  Riedel]{logan-iv-etal-2022-cutting}
Robert Logan~IV, Ivana Balazevic, Eric Wallace, Fabio Petroni, Sameer Singh,
  and Sebastian Riedel.
\newblock Cutting down on prompts and parameters: Simple few-shot learning with
  language models.
\newblock In \emph{Findings of the Association for Computational Linguistics:
  ACL 2022}, pp.\  2824--2835, Dublin, Ireland, May 2022. Association for
  Computational Linguistics.
\newblock \doi{10.18653/v1/2022.findings-acl.222}.
\newblock URL \url{https://aclanthology.org/2022.findings-acl.222}.

\bibitem[Lu et~al.(2021)Lu, Hernandez~Abrego, Ma, Ni, and
  Yang]{lu-etal-2021-multi}
Jing Lu, Gustavo Hernandez~Abrego, Ji~Ma, Jianmo Ni, and Yinfei Yang.
\newblock Multi-stage training with improved negative contrast for neural
  passage retrieval.
\newblock In \emph{Proceedings of the 2021 Conference on Empirical Methods in
  Natural Language Processing}, pp.\  6091--6103, Online and Punta Cana,
  Dominican Republic, November 2021. Association for Computational Linguistics.
\newblock \doi{10.18653/v1/2021.emnlp-main.492}.
\newblock URL \url{https://aclanthology.org/2021.emnlp-main.492}.

\bibitem[Luan et~al.(2021)Luan, Eisenstein, Toutanova, and
  Collins]{luan-2021-sparse}
Yi~Luan, Jacob Eisenstein, Kristina Toutanova, and Michael Collins.
\newblock {Sparse, Dense, and Attentional Representations for Text Retrieval}.
\newblock \emph{Transactions of the Association for Computational Linguistics},
  9:\penalty0 329--345, 04 2021.
\newblock ISSN 2307-387X.
\newblock \doi{10.1162/tacl_a_00369}.
\newblock URL \url{https://doi.org/10.1162/tacl\_a\_00369}.

\bibitem[Ma et~al.(2021)Ma, Korotkov, Yang, Hall, and
  McDonald]{ma-etal-2021-zero}
Ji~Ma, Ivan Korotkov, Yinfei Yang, Keith Hall, and Ryan McDonald.
\newblock Zero-shot neural passage retrieval via domain-targeted synthetic
  question generation.
\newblock In \emph{Proceedings of the 16th Conference of the European Chapter
  of the Association for Computational Linguistics: Main Volume}, pp.\
  1075--1088, Online, April 2021. Association for Computational Linguistics.
\newblock \doi{10.18653/v1/2021.eacl-main.92}.
\newblock URL \url{https://aclanthology.org/2021.eacl-main.92}.

\bibitem[Maia et~al.(2018)Maia, Handschuh, Freitas, Davis, McDermott, Zarrouk,
  and Balahur]{maia201818}
Macedo Maia, Siegfried Handschuh, Andr{\'e} Freitas, Brian Davis, Ross
  McDermott, Manel Zarrouk, and Alexandra Balahur.
\newblock Www'18 open challenge: financial opinion mining and question
  answering.
\newblock In \emph{Companion proceedings of the the web conference 2018}, pp.\
  1941--1942, 2018.
\newblock URL \url{https://doi.org/10.1145/3184558.3192301}.

\bibitem[McDonald et~al.(2018)McDonald, Brokos, and
  Androutsopoulos]{DeepRelevenceRanking}
Ryan McDonald, George Brokos, and Ion Androutsopoulos.
\newblock Deep relevance ranking using enhanced document-query interactions.
\newblock In \emph{Proceedings of the 2018 Conference on Empirical Methods in
  Natural Language Processing}, pp.\  1849--1860, Brussels, Belgium,
  October-November 2018. Association for Computational Linguistics.
\newblock \doi{10.18653/v1/D18-1211}.
\newblock URL \url{https://www.aclweb.org/anthology/D18-1211}.

\bibitem[Neelakantan et~al.(2022)Neelakantan, Xu, Puri, Radford, Han, Tworek,
  Yuan, Tezak, Kim, Hallacy, Heidecke, Shyam, Power, Nekoul, Sastry, Krueger,
  Schnurr, Such, Hsu, Thompson, Khan, Sherbakov, Jang, Welinder, and
  Weng]{Neelakantan2022TextAC}
Arvind Neelakantan, Tao Xu, Raul Puri, Alec Radford, Jesse~Michael Han, Jerry
  Tworek, Qiming Yuan, Nikolas~A. Tezak, Jong~Wook Kim, Chris Hallacy, Johannes
  Heidecke, Pranav Shyam, Boris Power, Tyna~Eloundou Nekoul, Girish Sastry,
  Gretchen Krueger, David~P. Schnurr, Felipe~Petroski Such, Kenny Sai-Kin Hsu,
  Madeleine Thompson, Tabarak Khan, Toki Sherbakov, Joanne Jang, Peter
  Welinder, and Lilian Weng.
\newblock Text and code embeddings by contrastive pre-training.
\newblock \emph{ArXiv}, abs/2201.10005, 2022.
\newblock URL \url{https://arxiv.org/abs/2201.10005}.

\bibitem[Nguyen et~al.(2016)Nguyen, Rosenberg, Song, Gao, Tiwary, Majumder, and
  Deng]{nguyen2016msmarco}
Tri Nguyen, Mir Rosenberg, Xia Song, Jianfeng Gao, Saurabh Tiwary, Rangan
  Majumder, and Li~Deng.
\newblock Ms marco: A human generated machine reading comprehension dataset.
\newblock In \emph{CoCo@NIPS}, 2016.
\newblock URL \url{http://ceur-ws.org/Vol-1773/CoCoNIPS_2016_paper9.pdf}.

\bibitem[Ni et~al.(2021)Ni, Qu, Lu, Dai, {\'{A}}brego, Ma, Zhao, Luan, Hall,
  Chang, and Yang]{DBLP:journals/corr/abs-2112-07899}
Jianmo Ni, Chen Qu, Jing Lu, Zhuyun Dai, Gustavo~Hern{\'{a}}ndez {\'{A}}brego,
  Ji~Ma, Vincent~Y. Zhao, Yi~Luan, Keith~B. Hall, Ming{-}Wei Chang, and Yinfei
  Yang.
\newblock Large dual encoders are generalizable retrievers.
\newblock \emph{CoRR}, abs/2112.07899, 2021.
\newblock URL \url{https://arxiv.org/abs/2112.07899}.

\bibitem[Nogueira \& Cho(2019)Nogueira and Cho]{nogueira2019passage}
Rodrigo Nogueira and Kyunghyun Cho.
\newblock Passage re-ranking with bert.
\newblock \emph{arXiv}, 2019.
\newblock URL \url{https://arxiv.org/abs/1901.04085}.

\bibitem[Nogueira et~al.(2020)Nogueira, Jiang, Pradeep, and
  Lin]{Nogueira20monot5}
Rodrigo Nogueira, Zhiying Jiang, Ronak Pradeep, and Jimmy Lin.
\newblock Document ranking with a pretrained sequence-to-sequence model.
\newblock In Trevor Cohn, Yulan He, and Yang Liu (eds.), \emph{Findings of the
  Association for Computational Linguistics: {EMNLP} 2020, Online Event, 16-20
  November 2020}, volume {EMNLP} 2020 of \emph{Findings of {ACL}}, pp.\
  708--718. Association for Computational Linguistics, 2020.
\newblock \doi{10.18653/v1/2020.findings-emnlp.63}.
\newblock URL \url{https://doi.org/10.18653/v1/2020.findings-emnlp.63}.

\bibitem[Oguz et~al.(2022)Oguz, Lakhotia, Gupta, Lewis, Karpukhin, Piktus,
  Chen, Riedel, Yih, Gupta, and Mehdad]{oguz-etal-2022-domain}
Barlas Oguz, Kushal Lakhotia, Anchit Gupta, Patrick Lewis, Vladimir Karpukhin,
  Aleksandra Piktus, Xilun Chen, Sebastian Riedel, Scott Yih, Sonal Gupta, and
  Yashar Mehdad.
\newblock Domain-matched pre-training tasks for dense retrieval.
\newblock In \emph{Findings of the Association for Computational Linguistics:
  NAACL 2022}, pp.\  1524--1534, Seattle, United States, July 2022. Association
  for Computational Linguistics.
\newblock \doi{10.18653/v1/2022.findings-naacl.114}.
\newblock URL \url{https://aclanthology.org/2022.findings-naacl.114}.

\bibitem[Palangi et~al.(2016)Palangi, Deng, Shen, Gao, He, Chen, Song, and
  Ward]{palangi2016deep}
Hamid Palangi, Li~Deng, Yelong Shen, Jianfeng Gao, Xiaodong He, Jianshu Chen,
  Xinying Song, and Rabab Ward.
\newblock Deep sentence embedding using long short-term memory networks:
  Analysis and application to information retrieval.
\newblock \emph{IEEE/ACM Transactions on Audio, Speech, and Language
  Processing}, 24\penalty0 (4):\penalty0 694–707, 2016.
\newblock URL \url{https://doi.org/10.1109/TASLP.2016.2520371}.

\bibitem[Qu et~al.(2021)Qu, Ding, Liu, Liu, Ren, Zhao, Dong, Wu, and
  Wang]{rocketqa}
Yingqi Qu, Yuchen Ding, Jing Liu, Kai Liu, Ruiyang Ren, Wayne~Xin Zhao, Daxiang
  Dong, Hua Wu, and Haifeng Wang.
\newblock {R}ocket{QA}: An optimized training approach to dense passage
  retrieval for open-domain question answering.
\newblock In \emph{Proceedings of the 2021 Conference of the North American
  Chapter of the Association for Computational Linguistics: Human Language
  Technologies}, pp.\  5835--5847, June 2021.
\newblock \doi{10.18653/v1/2021.naacl-main.466}.
\newblock URL \url{https://aclanthology.org/2021.naacl-main.466}.

\bibitem[Raffel et~al.(2020)Raffel, Shazeer, Roberts, Lee, Narang, Matena,
  Zhou, Li, and Liu]{2020t5}
Colin Raffel, Noam~M. Shazeer, Adam Roberts, Katherine Lee, Sharan Narang,
  Michael Matena, Yanqi Zhou, W.~Li, and Peter~J. Liu.
\newblock Exploring the limits of transfer learning with a unified text-to-text
  transformer.
\newblock \emph{Journal of Machine Learning Research}, 21/140:\penalty0 1--67,
  2020.
\newblock URL \url{http://jmlr.org/papers/v21/20-074.html}.

\bibitem[Reddi et~al.(2021)Reddi, Pasumarthi, Menon, Rawat, Yu, Kim, Veit, and
  Kumar]{DBLP:conf/aistats/ReddiPMRYKVK21}
Sashank~J. Reddi, Rama~Kumar Pasumarthi, Aditya~Krishna Menon, Ankit~Singh
  Rawat, Felix~X. Yu, Seungyeon Kim, Andreas Veit, and Sanjiv Kumar.
\newblock Rankdistil: Knowledge distillation for ranking.
\newblock In \emph{AISTATS}, pp.\  2368--2376, 2021.
\newblock URL \url{http://proceedings.mlr.press/v130/reddi21a.html}.

\bibitem[Ren et~al.(2021{\natexlab{a}})Ren, Lv, Qu, Liu, Zhao, She, Wu, Wang,
  and Wen]{ren-etal-2021-pair}
Ruiyang Ren, Shangwen Lv, Yingqi Qu, Jing Liu, Wayne~Xin Zhao, QiaoQiao She,
  Hua Wu, Haifeng Wang, and Ji-Rong Wen.
\newblock {PAIR}: Leveraging passage-centric similarity relation for improving
  dense passage retrieval.
\newblock In \emph{Findings of the Association for Computational Linguistics:
  ACL-IJCNLP 2021}, pp.\  2173--2183, Online, August 2021{\natexlab{a}}.
  Association for Computational Linguistics.
\newblock \doi{10.18653/v1/2021.findings-acl.191}.
\newblock URL \url{https://aclanthology.org/2021.findings-acl.191}.

\bibitem[Ren et~al.(2021{\natexlab{b}})Ren, Qu, Liu, Zhao, She, Wu, Wang, and
  Wen]{ren-etal-2021-rocketqav2}
Ruiyang Ren, Yingqi Qu, Jing Liu, Wayne~Xin Zhao, QiaoQiao She, Hua Wu, Haifeng
  Wang, and Ji-Rong Wen.
\newblock {R}ocket{QA}v2: A joint training method for dense passage retrieval
  and passage re-ranking.
\newblock In \emph{Proceedings of the 2021 Conference on Empirical Methods in
  Natural Language Processing}, pp.\  2825--2835, Online and Punta Cana,
  Dominican Republic, November 2021{\natexlab{b}}. Association for
  Computational Linguistics.
\newblock \doi{10.18653/v1/2021.emnlp-main.224}.
\newblock URL \url{https://aclanthology.org/2021.emnlp-main.224}.

\bibitem[Ren et~al.(2022)Ren, Qu, Liu, Zhao, Wu, Ding, Wu, Wang, and
  Wen]{ren2022dreval}
Ruiyang Ren, Yingqi Qu, Jing Liu, Wayne~Xin Zhao, Qifei Wu, Yuchen Ding, Hua
  Wu, Haifeng Wang, and Ji-Rong Wen.
\newblock A thorough examination on zero-shot dense retrieval, 2022.
\newblock URL \url{https://arxiv.org/abs/2204.12755}.

\bibitem[Rosa et~al.(2022)Rosa, Bonifacio, Jeronymo, Abonizio, Fadaee, Lotufo,
  and Nogueira]{rosa2022no}
Guilherme~Moraes Rosa, Luiz Bonifacio, Vitor Jeronymo, Hugo Abonizio, Marzieh
  Fadaee, Roberto Lotufo, and Rodrigo Nogueira.
\newblock No parameter left behind: How distillation and model size affect
  zero-shot retrieval.
\newblock \emph{arXiv preprint arXiv:2206.02873}, 2022.

\bibitem[Sachan et~al.(2022)Sachan, Lewis, Joshi, Aghajanyan, Yih, Pineau, and
  Zettlemoyer]{sachan2022improving}
Devendra~Singh Sachan, Mike Lewis, Mandar Joshi, Armen Aghajanyan, Wen-tau Yih,
  Joelle Pineau, and Luke Zettlemoyer.
\newblock Improving passage retrieval with zero-shot question generation.
\newblock \emph{arXiv}, 2022.
\newblock URL \url{https://arxiv.org/abs/2204.07496}.

\bibitem[Sanh et~al.(2021)Sanh, Webson, Raffel, Bach, Sutawika, Alyafeai,
  Chaffin, Stiegler, Scao, Raja, et~al.]{sanh2021tzero}
Victor Sanh, Albert Webson, Colin Raffel, Stephen~H Bach, Lintang Sutawika,
  Zaid Alyafeai, Antoine Chaffin, Arnaud Stiegler, Teven~Le Scao, Arun Raja,
  et~al.
\newblock Multitask prompted training enables zero-shot task generalization.
\newblock \emph{arXiv preprint arXiv:2110.08207}, 2021.

\bibitem[Santhanam et~al.(2022)Santhanam, Khattab, Saad-Falcon, Potts, and
  Zaharia]{Santhanam2022ColBERTv2EA}
Keshav Santhanam, Omar Khattab, Jon Saad-Falcon, Christopher Potts, and Matei
  Zaharia.
\newblock {C}ol{BERT}v2: Effective and efficient retrieval via lightweight late
  interaction.
\newblock In \emph{Proceedings of the 2022 Conference of the North American
  Chapter of the Association for Computational Linguistics: Human Language
  Technologies}, pp.\  3715--3734, July 2022.
\newblock URL \url{https://aclanthology.org/2022.naacl-main.272}.

\bibitem[Schick \& Sch{\"u}tze(2021{\natexlab{a}})Schick and
  Sch{\"u}tze]{schick-schutze-2021-exploiting}
Timo Schick and Hinrich Sch{\"u}tze.
\newblock Exploiting cloze-questions for few-shot text classification and
  natural language inference.
\newblock In \emph{Proceedings of the 16th Conference of the European Chapter
  of the Association for Computational Linguistics: Main Volume}, pp.\
  255--269, Online, April 2021{\natexlab{a}}. Association for Computational
  Linguistics.
\newblock \doi{10.18653/v1/2021.eacl-main.20}.
\newblock URL \url{https://aclanthology.org/2021.eacl-main.20}.

\bibitem[Schick \& Sch{\"u}tze(2021{\natexlab{b}})Schick and
  Sch{\"u}tze]{schick-schutze-2021-just}
Timo Schick and Hinrich Sch{\"u}tze.
\newblock It{'}s not just size that matters: Small language models are also
  few-shot learners.
\newblock In \emph{Proceedings of the 2021 Conference of the North American
  Chapter of the Association for Computational Linguistics: Human Language
  Technologies}, pp.\  2339--2352, Online, June 2021{\natexlab{b}}. Association
  for Computational Linguistics.
\newblock \doi{10.18653/v1/2021.naacl-main.185}.
\newblock URL \url{https://aclanthology.org/2021.naacl-main.185}.

\bibitem[Schick \& Sch{\"u}tze(2021{\natexlab{c}})Schick and
  Sch{\"u}tze]{schick-schutze-2021-shot}
Timo Schick and Hinrich Sch{\"u}tze.
\newblock Few-shot text generation with natural language instructions.
\newblock In \emph{Proceedings of the 2021 Conference on Empirical Methods in
  Natural Language Processing}, pp.\  390--402, Online and Punta Cana,
  Dominican Republic, November 2021{\natexlab{c}}. Association for
  Computational Linguistics.
\newblock \doi{10.18653/v1/2021.emnlp-main.32}.
\newblock URL \url{https://aclanthology.org/2021.emnlp-main.32}.

\bibitem[Shakeri et~al.(2020)Shakeri, Nogueira~dos Santos, Zhu, Ng, Nan, Wang,
  Nallapati, and Xiang]{siamak20}
Siamak Shakeri, Cicero Nogueira~dos Santos, Henghui Zhu, Patrick Ng, Feng Nan,
  Zhiguo Wang, Ramesh Nallapati, and Bing Xiang.
\newblock End-to-end synthetic data generation for domain adaptation of
  question answering systems.
\newblock In \emph{Proceedings of the 2020 Conference on Empirical Methods in
  Natural Language Processing (EMNLP)}, pp.\  5445--5460. Association for
  Computational Linguistics, 2020.
\newblock URL \url{https://aclanthology.org/2020.emnlp-main.439}.

\bibitem[Thakur et~al.(2021)Thakur, Reimers, R{\"u}ckl{\'e}, Srivastava, and
  Gurevych]{thakur2021beir}
Nandan Thakur, Nils Reimers, Andreas R{\"u}ckl{\'e}, Abhishek Srivastava, and
  Iryna Gurevych.
\newblock {BEIR}: A heterogeneous benchmark for zero-shot evaluation of
  information retrieval models.
\newblock In \emph{Thirty-fifth Conference on Neural Information Processing
  Systems Datasets and Benchmarks Track (Round 2)}, 2021.
\newblock URL \url{https://openreview.net/forum?id=wCu6T5xFjeJ}.

\bibitem[Thoppilan et~al.(2022)Thoppilan, Freitas, Hall, Shazeer, Kulshreshtha,
  Cheng, Jin, Bos, Baker, Du, Li, Lee, Zheng, Ghafouri, Menegali, Huang,
  Krikun, Lepikhin, Qin, Chen, Xu, Chen, Roberts, Bosma, Zhou, Chang, Krivokon,
  Rusch, Pickett, Meier{-}Hellstern, Morris, Doshi, Santos, Duke, Soraker,
  Zevenbergen, Prabhakaran, Diaz, Hutchinson, Olson, Molina, Hoffman{-}John,
  Lee, Aroyo, Rajakumar, Butryna, Lamm, Kuzmina, Fenton, Cohen, Bernstein,
  Kurzweil, Aguera{-}Arcas, Cui, Croak, Chi, and Le]{lamda}
Romal Thoppilan, Daniel~De Freitas, Jamie Hall, Noam Shazeer, Apoorv
  Kulshreshtha, Heng{-}Tze Cheng, Alicia Jin, Taylor Bos, Leslie Baker, Yu~Du,
  YaGuang Li, Hongrae Lee, Huaixiu~Steven Zheng, Amin Ghafouri, Marcelo
  Menegali, Yanping Huang, Maxim Krikun, Dmitry Lepikhin, James Qin, Dehao
  Chen, Yuanzhong Xu, Zhifeng Chen, Adam Roberts, Maarten Bosma, Yanqi Zhou,
  Chung{-}Ching Chang, Igor Krivokon, Will Rusch, Marc Pickett, Kathleen~S.
  Meier{-}Hellstern, Meredith~Ringel Morris, Tulsee Doshi, Renelito~Delos
  Santos, Toju Duke, Johnny Soraker, Ben Zevenbergen, Vinodkumar Prabhakaran,
  Mark Diaz, Ben Hutchinson, Kristen Olson, Alejandra Molina, Erin
  Hoffman{-}John, Josh Lee, Lora Aroyo, Ravi Rajakumar, Alena Butryna, Matthew
  Lamm, Viktoriya Kuzmina, Joe Fenton, Aaron Cohen, Rachel Bernstein, Ray
  Kurzweil, Blaise Aguera{-}Arcas, Claire Cui, Marian Croak, Ed~H. Chi, and
  Quoc Le.
\newblock Lamda: Language models for dialog applications.
\newblock \emph{CoRR}, abs/2201.08239, 2022.
\newblock URL \url{https://arxiv.org/abs/2201.08239}.

\bibitem[Thorne et~al.(2018)Thorne, Vlachos, Christodoulopoulos, and
  Mittal]{thorne-etal-2018-fever}
James Thorne, Andreas Vlachos, Christos Christodoulopoulos, and Arpit Mittal.
\newblock {FEVER}: a large-scale dataset for fact extraction and
  {VER}ification.
\newblock In \emph{Proceedings of the 2018 Conference of the North {A}merican
  Chapter of the Association for Computational Linguistics: Human Language
  Technologies, Volume 1 (Long Papers)}, pp.\  809--819, New Orleans,
  Louisiana, June 2018. Association for Computational Linguistics.
\newblock \doi{10.18653/v1/N18-1074}.
\newblock URL \url{https://aclanthology.org/N18-1074}.

\bibitem[Wang et~al.(2022)Wang, Thakur, Reimers, and
  Gurevych]{wang-etal-2022-gpl}
Kexin Wang, Nandan Thakur, Nils Reimers, and Iryna Gurevych.
\newblock {GPL}: Generative pseudo labeling for unsupervised domain adaptation
  of dense retrieval.
\newblock In \emph{Proceedings of the 2022 Conference of the North American
  Chapter of the Association for Computational Linguistics: Human Language
  Technologies}, pp.\  2345--2360, Seattle, United States, July 2022.
  Association for Computational Linguistics.
\newblock \doi{10.18653/v1/2022.naacl-main.168}.
\newblock URL \url{https://aclanthology.org/2022.naacl-main.168}.

\bibitem[Wei et~al.(2022{\natexlab{a}})Wei, Bosma, Zhao, Guu, Yu, Lester, Du,
  Dai, and Le]{flan}
Jason Wei, Maarten~Paul Bosma, Vincent Zhao, Kelvin Guu, Adams~Wei Yu, Brian
  Lester, Nan Du, Andrew~Mingbo Dai, and Quoc~V. Le.
\newblock Finetuned language models are zero-shot learners.
\newblock In \emph{International Conference on Learning Representations},
  2022{\natexlab{a}}.
\newblock URL \url{https://openreview.net/forum?id=gEZrGCozdqR}.

\bibitem[Wei et~al.(2022{\natexlab{b}})Wei, Tay, Bommasani, Raffel, Zoph,
  Borgeaud, Yogatama, Bosma, Zhou, Metzler, Chi, Hashimoto, Vinyals, Liang,
  Dean, and Fedus]{wei-2022-emergent}
Jason Wei, Yi~Tay, Rishi Bommasani, Colin Raffel, Barret Zoph, Sebastian
  Borgeaud, Dani Yogatama, Maarten Bosma, Denny Zhou, Donald Metzler, Ed~H.
  Chi, Tatsunori Hashimoto, Oriol Vinyals, Percy Liang, Jeff Dean, and William
  Fedus.
\newblock Emergent abilities of large language models.
\newblock \emph{Transactions on Machine Learning Research}, 2022{\natexlab{b}}.
\newblock URL \url{https://openreview.net/forum?id=yzkSU5zdwD}.

\bibitem[Wu et~al.(2019)Wu, Guo, Simcha, Dopson, and Kumar]{wu2019efficient}
Xiang Wu, Ruiqi Guo, David Simcha, Dave Dopson, and Sanjiv Kumar.
\newblock Efficient inner product approximation in hybrid spaces.
\newblock \emph{arXiv}, 2019.
\newblock URL \url{https://arxiv.org/abs/1903.08690}.

\bibitem[Xiong et~al.(2017)Xiong, Dai, Callan, Liu, and Power]{xiong2017end}
Chenyan Xiong, Zhuyun Dai, Jamie Callan, Zhiyuan Liu, and Russell Power.
\newblock End-to-end neural ad-hoc ranking with kernel pooling.
\newblock In \emph{Proceedings of the 40th International ACM SIGIR Conference
  on Research and Development in Information Retrieval}, SIGIR '17, pp.\
  55–64, New York, NY, USA, 2017.
\newblock URL \url{https://doi.org/10.1145/3077136.3080809}.

\bibitem[Xiong et~al.(2021)Xiong, Xiong, Li, Tang, Liu, Bennett, Ahmed, and
  Overwijk]{ance}
Lee Xiong, Chenyan Xiong, Ye~Li, Kwok-Fung Tang, Jialin Liu, Paul~N. Bennett,
  Junaid Ahmed, and Arnold Overwijk.
\newblock Approximate nearest neighbor negative contrastive learning for dense
  text retrieval.
\newblock In \emph{International Conference on Learning Representations}, 2021.
\newblock URL \url{https://openreview.net/forum?id=zeFrfgyZln}.

\bibitem[Yang et~al.(2018)Yang, Qi, Zhang, Bengio, Cohen, Salakhutdinov, and
  Manning]{yang-etal-2018-hotpotqa}
Zhilin Yang, Peng Qi, Saizheng Zhang, Yoshua Bengio, William Cohen, Ruslan
  Salakhutdinov, and Christopher~D. Manning.
\newblock {H}otpot{QA}: A dataset for diverse, explainable multi-hop question
  answering.
\newblock In \emph{Proceedings of the 2018 Conference on Empirical Methods in
  Natural Language Processing}, pp.\  2369--2380, Brussels, Belgium,
  October-November 2018. Association for Computational Linguistics.
\newblock \doi{10.18653/v1/D18-1259}.
\newblock URL \url{https://aclanthology.org/D18-1259}.

\bibitem[Yih et~al.(2011)Yih, Toutanova, Platt, and
  Meek]{yih-etal-2011-learning}
Wen-tau Yih, Kristina Toutanova, John~C. Platt, and Christopher Meek.
\newblock Learning discriminative projections for text similarity measures.
\newblock In \emph{Proceedings of the Fifteenth Conference on Computational
  Natural Language Learning}, pp.\  247--256, Portland, Oregon, USA, June 2011.
  Association for Computational Linguistics.
\newblock URL \url{https://aclanthology.org/W11-0329}.

\bibitem[Zhang et~al.(2022)Zhang, Gong, Shen, Lv, Duan, and
  Chen]{zhang2022adversarial}
Hang Zhang, Yeyun Gong, Yelong Shen, Jiancheng Lv, Nan Duan, and Weizhu Chen.
\newblock Adversarial retriever-ranker for dense text retrieval.
\newblock In \emph{International Conference on Learning Representations}, 2022.
\newblock URL \url{https://openreview.net/forum?id=MR7XubKUFB}.

\end{thebibliography}
\bibliographystyle{iclr2023_conference}

\appendix

\section{Analysis on Prompts}
\label{app:analysis_on_prompts}
Table \ref{tab:promt_template} shows the list of prompt templates on different BEIR datasets. In order to further analysis the difference between zero-shot and few-shot prompts, we compare the few-shot and zero-shot generated queries given the same paragraph, randomly sampled from three datasets in Table \ref{tab:examples}. We observe that in general, the few-shot generated queries are closer to the original queries, while zero-shot queries are mostly questions. For example, in the ArguAna dataset, the few-shot queries are in general longer and more claim-like. In contrary, the zero-shot queries are most short question-like queries. Interestingly, for the HotpotQA dataset, even though both few-shot and zero-shot queries are generating questions-like queries, few-shot queries sometimes generate multi-hop questions, while zero-shot mostly generates single-hop questions.
We further conduct first word distribution across different generation models for all datasets in
Figure~\ref{fig:topwords_appendix}.

\begin{table*}[h]%
\centering
\footnotesize
\begin{tabular}{p{3cm}|p{10cm}}
\toprule
 Dataset & Prompt  \\
\midrule
ArguAna
 &
\nlp{0 Argument: {passage} X 1 Counter argument: {query} X}
\\ 
\midrule
FiQA
 &
\nlp{0 {passage} X 1 {query} X}
 \\ 
 \midrule
HotpotQA
&
\nlp{0 Evidence: {passage} X 1 Vexed question: {query} X} 
 \\ 
 \midrule
DBPedia-Entity
&
\nlp{0 entity: {passage} X 1 query: {query} X}
 \\ 
  \midrule
NFCorpus
&
\nlp{0 Article: {passage} X 1 Query: {query} X}
\\ 
  \midrule
Touché-2020
&
\nlp{0 {passage} X 1 Debate: {query} X} 
\\ 
   \midrule
  TREC-Covid
&
\nlp{0 {passage} X 1 Question: {query} X}
 \\ 
    \midrule
SciFact
&
\nlp{0  {passage} X 1 Finding: {query} X}
\\ 
    \midrule
SCIDOCS
&
\nlp{0  {passage} X 1 The passage is about {query} X}
\\ 
    \midrule
FEVER
&
\nlp{0 {passage} X 1 Is it true that {query} X}
\\ 
\bottomrule
\end{tabular}
\caption{\label{tab:promt_template} Prompt template for each dataset.}
\vspace{-1mm}
\end{table*}

\begin{table*}%
\centering
\footnotesize
\begin{tabular}{p{4.25cm}|p{3.5cm}|p{3.85cm}|p{2.7cm}}
\toprule
 Paragraph & Few-shot & Zero-shot &  Analysis  \\
\midrule
ArguAna: Banning skin whitening on such a basis also requires the acceptance of the racial overtones. Some form of tan is popular in many societies of people of European ethnic origins – that is not a racial matter, it is more based on economic social perceptions (that of holidays to warmer climates). Ascribing a racial element to everything to do with skin tone is at best a lazy analysis.  Irrespective of issues of race and perceptions of ethnic origins, and its intersection with beauty standards, some people will be given advantages in life due to their appearance. Banning a certain form of cosmetic, even if it can have some racial and ethnic undertones, won’t change that.
 &
Query 1: Banning skin whitening on such a basis also requires the acceptance of the racial overtones. However, the racial overtones are simply a side effect of the product being used incorrectly in order to attain a lighter skin tone than they are naturally born with. The product itself is not racist, but rather the people who use it. For example, a person of Asian descent may use a skin whitening cream to attain a lighter skin tone so that they may fit into a more white-dominant society. The cream itself is not racist, but rather the fact that the person has used it to attain a lighter skin tone so that they may fit into a more white-dominant society.
&
Query 1: What did they try to ban? \newline
Query 2: Is skin whitening racist? \newline
Query 3: Banning skin whitening on such a basis also requires the acceptance of the racial overtones. Some form of tan is popular in many societies of people of European ethnic origins – that is not a racial matter, it is more based on economic social perceptions (that of holidays to warmer climates). \newline
Query 4: What are some of the racial overtones in skin-whitening? \newline
Query 5: Are tans a sign of affluence? 
 & In ArguAna dataset, the few-shot examples are more statement like and are longer than zero-shot. 
\\ 
\midrule
Touché-2020: Psychologically,diet-wise and academically. Housewives can be both educated and uneducated but the fact that they spend more time with their children means that they put in a greater effort and since they are not preoccupied with their job(read: cellphone); they pay the close attention needed to keep their children out of trouble. Think of housewives as inspectors in your home...
 &
Query 1: Housewives are better parents
 \newline
Query 2: Should housewives be considered better parents?
 \newline
&
Query 1: ,diet-wise and academically.
 \newline
Query 2: People want to be relaxed all the time(At home AND at work) that is why all the research on work-life balance.
 \newline
Query 3:has shown that home-makers are more relaxed, less stressed, healthier, happier - the list goes on.
 \newline
& The few-shot generates argument-like queries that are more controversial, while zero-shot generate random statements that sometimes even have grammar problems.

 \\ 
 \midrule
HotpotQA: The Yung See San Fong House, or \"Young's Home in the Heart of the Hills\", is a house in Los Gatos in the U.S. state of California. It is a combination of Asian decorative motifs and pagoda roofs together with Western massing and layout. It was completed in 1917 for Ruth Comfort Mitchell Young, a writer, and her husband, Sanborn Young, a gentleman farmer, conservationist and later California State Senator.
&
Query 1: \textcolor{blue}{Which author was the husband of Ruth Comfort Mitchell Young?} \newline
Query 2: Which is known as the Yount-Lee House \newline
Query 3: Who is the architect of the Yung See San Fong House? \newline
Query 4: The Yung See San Fong House, or \"Young's Home in the Heart of the Hills\", is a house in which U.S. state? 

& 
Query 1: 
Yung See San Fong House was completed in 1917 for Ruth Comfort Mitchell Young, a writer, and her husband, Sanborn Young. \newline
Query 2: How old is the Yung See San Fong House? \newline
Query 3: Yung See San Fong House was completed in 1917 for Ruth Comfort Mitchell Young, a writer, and her husband, Sanborn Young. 

& Few-shot examples sometimes will create multihop questions indicated in \textcolor{blue}{blue}, which rarely happens in zero-shot examples. 
 \\ 

\bottomrule
\end{tabular}
\caption{\label{tab:examples} Few-shot and zero-shot generated queries randomly sampled from ArguAna, FiQA and HotpotQA dataset.}
\vspace{-1mm}
\end{table*}

\input{tables_and_figures/analysis_top_first_unigram_appendix}

\section{Detailed Implementation}
\begin{figure}[h]
    \centering
    \includegraphics[width=\linewidth]{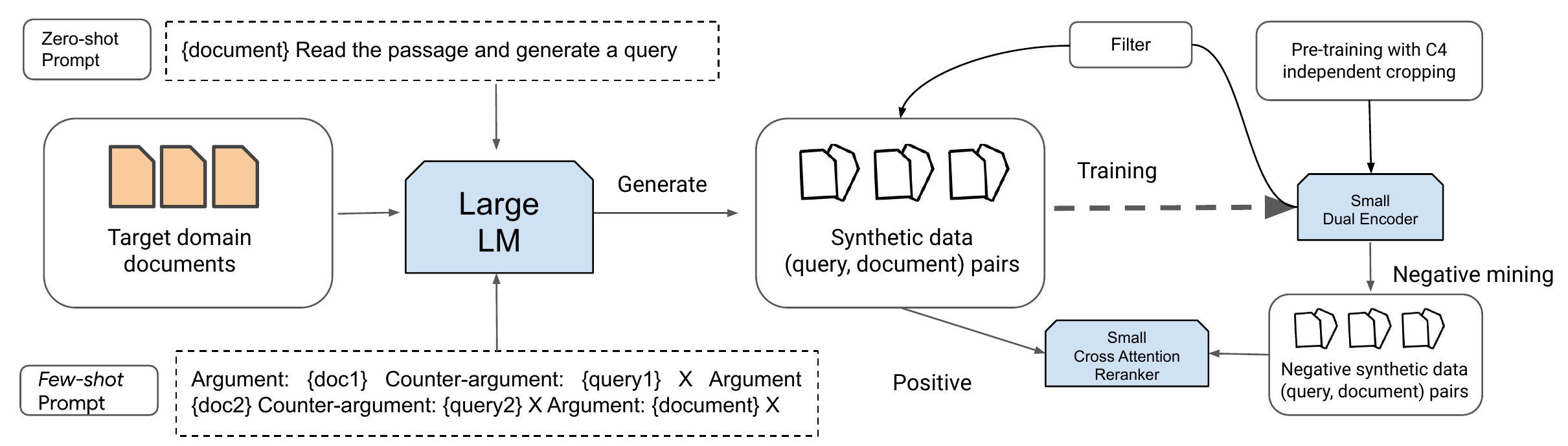}
     \caption{\namep Training pipeline.}
     \label{fig:overall}
\end{figure}

Figure \ref{fig:overall} shows the overall process of \namep,  the details of which are in Section \ref{sec:model}.

\section{Query Generation Statistics}

\begin{wraptable}{R}{6cm}
\resizebox{6cm}{!}{
\footnotesize
  \begin{tabular}{lccc}
  \toprule
               & Few-shot  & Zero-shot  & NQ QGen 
               \\ \midrule 
ArguAna        & 98.2  & 26.0 & 9.7                         \\
Touché-2020    & 7.8  & 13.4 & 9.8                         \\
TREC-Covid     & 10.8  & 11.4 & 10.2                         \\
NFCorpus      & 8.3  & 11.5 & 10.3                        \\
HotpotQA       & 11.2  & 12.2 & 8.8                         \\
DBPedia-Entity & 8.2  & 13.8 & 8.8                         \\
Fever          & 12.1  & 10.7 & 8.8                         \\
Climate-Fever  & 12.9  & 10.7 & 8.8                         \\
SciFact        & 12.6  & 12.4 & 10.0                        \\
SCIDOCS        & 7.4  & 15.7 & 10.7                         \\
FiQA-2018      & 12.5  & 10.1 & 9.5                        \\

               \midrule 
 AVG. & 17.8 & 13.5  & 9.6 \\
 \bottomrule            
\end{tabular}
}
  \caption{Average query length.}%
  \label{tab:analysis_length}
\end{wraptable}

In Table~\ref{tab:analysis_length}, we analyze the length of the generated questions by different query generation systems. Note that NQ-QGen always generates short queries due to the query generation models being fine-tuned on the NQ dataset, and all of the generated questions have similar length to those questions of NQ. Interestingly, zero-shot \name already obtains more variance in terms of length compared to NQ-QGen. Finally, few-shot \name offers significantly more variance in terms of the length of generated queries.

\section{Author Contributions}

{\footnotesize
{\em Zhuyun Dai}: Propose few-shot retrieval idea. Perform early experiments to validate the approach.  Retrieval pipeline infrastructure. \name experiment.

{\em Vincent Zhao}: Propose few-shot retrieval idea. Main developer for the retrieval pipeline infrastructure. FLAN data generation. \name experiment.

{\em Ji Ma}: Advise research directions. Retrieval infrastructure design and implementation.  Propose round trip filtering.  NQ QGEN experiments. \name experiment.

{\em Yi Luan:} Conduct all \namep experiments. Analyzing \name generated data. Early distillation experiments.

{\em Jianmo Ni:} Main developer for transformer dual encoder and reranking modeling development. \name experiment.

{\em Jing Lu:} Early distillation experiments. Reranking modeling development.

{\em Anton Bakalov:} Reranking evaluation code support. Discussion.

{\em Kelvin Guu:} Early few-shot retrieval idea. Advise research directions.

{\em Keith B. Hall:} Advise research directions. Mentor researchers.  Prompt design and analysis.

{\em Ming-Wei Chang:} Project initiator. Team organization. Advise research directions. Mentor researchers. Prompt design and analysis.

}

\include{tables_and_figures/result_summary_bar}
\end{document}